\documentclass{amsart}
\usepackage{amssymb}
\usepackage{amsthm}
\usepackage{amsmath, mathtools, bm, bbm}
\usepackage{fullpage,textcomp}
\usepackage{multirow}
\usepackage[table,xcdraw]{xcolor}
\usepackage[export]{adjustbox}

\usepackage[singlelinecheck=true, center]{caption}
\usepackage[justification=centering, labelfont=bf]{subcaption}

\newtheorem{remark}{Remark}

\newcommand{\bb}[1]{\mathbf{#1}}
\newcommand{\nn}[1]{\left\|#1\right\|}
\newcommand{\mmu}{\bm{\mu}}
\newcommand{\IP}[3]{\left\langle #2, #3\right\rangle_{#1}}
\newcommand{\kdf}{\mathrm{ker}\,f'}

\DeclareMathOperator*{\argmin}{arg\,min}

\begin{document}

\title{Nonlinear Level Set Learning for Function Approximation on Sparse Data with Applications to Parametric Differential Equations}

\author{Anthony Gruber$^{1,*}$}
\author{Max Gunzburger$^1$}
\author{Lili Ju$^2$}
\author{Yuankai Teng$^2$}
\author{Zhu Wang$^2$}
\thanks{$^*$Corresponding author: (Anthony Gruber)  anthony.gruber@fsu.edu}

\email{agruber@fsu.edu, mgunzburger@fsu.edu, ju@math.sc.edu, yteng@email.sc.edu, wangzhu@math.sc.edu}

\address{$^1$ Department of Scientific Computing, Florida State University, 400 Dirac Science Library, Tallahassee, FL 32306, USA}

\address{$^2$ Department of Mathematics, University of South Carolina, 1523 Greene Street, Columbia, SC 29208, USA}

\begin{abstract}
A dimension reduction method based on the ``Nonlinear Level set Learning'' (NLL) approach is presented for the pointwise prediction of functions which have been sparsely sampled.  Leveraging geometric information provided by the Implicit Function Theorem, the proposed algorithm effectively reduces the input dimension to the theoretical lower bound with minor accuracy loss, providing a one-dimensional representation of the function which can be used for regression and sensitivity analysis.  Experiments and applications are presented which compare this modified NLL with the original NLL and the Active Subspaces (AS) method.  While accommodating sparse input data, the proposed algorithm is shown to train quickly and provide a much more accurate and informative reduction than either AS or the original NLL on two example functions with high-dimensional domains, as well as two state-dependent quantities depending on the solutions to parametric differential equations.

\vspace{0.5pc}


\end{abstract}

\maketitle

% \begin{keyword}
% %% keywords here, in the form: keyword \sep keyword
% Nonlinear level set learning \sep function approximation \sep sparse data \sep nonlinear dimensionality reduction
% %% PACS codes here, in the form: \PACS code \sep code
% % \PACS 0000 \sep 1111
% %% MSC codes here, in the form: \MSC code \sep code
% %% or \MSC[2008] code \sep code (2000 is the default)
% \MSC[2020] 65D15 \sep 65D40
% \end{keyword}


%%%% Start %%%%%%
\section{Introduction}
It is frequently the case that  scientists or engineers need to draw conclusions about the output of a function based on limited or incomplete data.  Such situations arise, for example, when the output depends on the solution of expensive  differential equations, or when lack of time and resources precludes the collection of sufficient high-quality samples.  When this occurs, it becomes critical to maximize the value of the limited resources at hand, which requires informed algorithms for dimension reduction.  

More specifically, let $U \subset \mathbb{R}^n$ be a bounded domain and consider the problem of approximating a continuously differentiable scalar function $f:U \to \mathbb{R}$ based on some predefined samples $\{\bb{x}^s, f(\bb{x}^s), \nabla f(\bb{x}^s)\}_{s\in S}$ of the function and its gradient vector field.  Note that $f$ may represent either a scalar quantity or some component of a vector quantity, so that no generality is lost with this consideration.  Additionally, let $\rho:\mathbb{R}^n \to \mathbb{R}^+$ be a probability density function supported on $U$ such that $f$ is square-integrable with respect to $\rho$, i.e. 
\[ \nn{f}_2^2 \coloneqq \int_U f(\bb{x})^2\rho(\bb{x})\,dx < \infty. \]
To generate a pointwise approximation to $f$, it is reasonable to seek a function $\tilde{f}:U\to\mathbb{R}$ which satisfies the minimization condition
\begin{equation}\label{eq:reg}
    \tilde{f}(\bb{x}) \in \argmin_{g\in C^1(U)} \nn{f(\bb{x}) - g(\bb{x})}_2^2,
\end{equation}
However, if the dimension $n$ is large relative to the number $|S|$ of available samples (i.e. sparse data), training a regression model to approximate $f$ directly becomes infeasible.  Indeed, unless the training data itself has a hidden low-dimensional structure, unsupervised learning methods such as  feed-forward neural networks are prone to overfitting, leading to poor accuracy on new data as a result of inadequate generalizability.  Therefore, it is necessary to employ some kind of dimension reduction to increase the density of the sampling data to the point where it is useful for approximating solutions to \eqref{eq:reg}.  

A prototypical example of this issue arises when studying the numerical solutions of differential equations with limited computational budget.  Let $I$ be a multi-index, $\bm{\beta} \in \mathbb{R}^m$, and consider a $k^{th}$-order parameterized system of $R\in\mathbb{N}$ partial differential equations (PDE) for the function $\bb{u}:\mathbb{R}^m \times \mathbb{R}^n \to \mathbb{R}^l$,
\begin{equation}\label{eq:pde}
F^r \left(\bm{\beta}, \bb{x},\bb{u}, \frac{\partial^{|I|}\bb{u}}{\partial x^I} \right) = 0, \qquad 1 \leq r \leq R, \qquad 1 \leq |I| \leq k,
\end{equation}
which may depend on some number of initial or boundary conditions.  Suppose additionally that the assignment $\bm\beta \mapsto \bb{u}(\bm{\beta}, \bb{x})$ is unique, so that solutions to \eqref{eq:pde} are parameterized by the variables $\bm{\beta}$. For prediction and sensitivity analysis it is often necessary to compute the value of some functional $\mathcal{K}(\bb{u})$ on PDE solutions $\bb{u}\in C^k(\mathbb{R}^m \times \mathbb{R}^n; \mathbb{R}^l)$ (e.g. temperature or total kinetic energy) which is implicitly a function of the parameters $\bm{\beta}$, i.e. $\mathcal{K}(\bm{\beta}) = \mathcal{K}(\bb{u}(\bm{\beta},\bb{x}))$.  On the other hand, it is usually not feasible to simulate the (potentially expensive) system \eqref{eq:pde} for every parameter configuration desired, so it is necessary to have a reasonable yet inexpensive approximation to $\mathcal{K}$ which can be computed for any $\bm{\beta}$ in place of numerically solving \eqref{eq:pde}.  In the language of before, this means finding $\tilde{\mathcal{K}}$ satisfying 
\[ \tilde{\mathcal{K}}(\bm{\beta}) \in \argmin_{\mathcal{G}:\mathbb{R}^m\to\mathbb{R}} \nn{\mathcal{K}(\bm{\beta}) - \mathcal{G}(\bm{\beta})}^2_2. \]
This poses difficulty when the set of configurations $\beta$ for which solutions are available is sparse in $\mathbb{R}^m$, as occurs when solutions to \eqref{eq:pde} require hours (or even days) on a supercomputer to obtain.  On the other hand, Section~\ref{sec:pde} shows that dimension reduction can be used to produce certain low-dimensional approximations to functionals like $\mathcal{K}$ which reliably approximate the true values.

% dimension reduction techniques are similarly important for obtaining an approximation $\tilde{\mathcal{K}}$ which is pointwise-close to the true solution. 

The state-of-the-art strategies for dimension reduction can be loosely categorized as either intrinsic or extrinsic based on how they organize the available data.  Intrinsic methods look for patterns directly within the input samples, which can be exploited for lower-dimensional classification and clustering once revealed.  At the present time, there are several effective dimension-reduction methods which operate intrinsically, including Isomap, Locally Linear Embedding, and others (see e.g. \cite{Isomap,LLE,bengio2003,budninskiy2019} and references therein).  Moreover, these methods have demonstrated good performance on low-dimensional encoding problems as well as the recovery of geodesic distances along the data manifold (see e.g.  \cite{choi2007, zhang2018, lee2004}).  On the other hand, intrinsic methods are of no use when the data is unstructured, since there is no low-dimensional structure to be found.  This motivates the search for extrinsic dimension-reduction methods, which do not assume the given data possesses any structure at all.  Instead, these methods take advantage of the structure which is inherited from an external object acting on the input space.  For example, consider a surjective $C^1$ mapping onto a low-dimensional target, which stratifies the source according to its level sets.  If this induced level set structure can be learned, then this notion can be used to produce a low-dimensional approximation to the original function.  The advantage of such extrinsic methods is that they are applicable to any kind of sampling, including sparsity patterns which are indistinguishable from random noise. Conversely, these techniques are necessarily dependent on the object the source data inherits from, meaning there can be no singular solution which applies to every notion of inherited structure.  

% Moreover, it can be difficult to know how to make use of an extrinsic approach when developing a viable dimension reduction method.

\begin{remark}
We use the notation $f'(\bb{x})$ to denote the derivative of $f$, i.e. the linear map induced by $f$  satisfying $f'(\bb{x})\bb{v} = \IP{}{\nabla f(\bb{x})}{\bb{v}}$ for all vectors $\bb{v}$.  Similarly, if $\{\bb{e}_1,...,\bb{e}_n\}$ denote the standard basis for $\mathbb{R}^n$ then we let $f_i \coloneqq f'(\bb{x})\bb{e}_i = \partial f(\bb{x})/ \partial x^i$ denote the derivative of $f$ with respect to $\bb{e}_i$.  The extension of this notation to vector-valued functions is straightforward.
\end{remark}

Thankfully, in the case of $C^1$  functions there is a good deal of information provided by classical theory which can be exploited for algorithm design.   Given a regular value $y_0\in\mathbb{R}$, the Implicit Function Theorem (IFT) guarantees that the level set $f^{-1}(y_0) \subset \mathbb{R}^n$ is a differentiable submanifold of $\mathbb{R}^n$, and around any preimage $\bb{x}_0\in f^{-1}(y_0)$ there is the representation
\[T_{\bb{x}_0}f^{-1}(y_0) = \kdf(\bb{x}_0),\]
which characterizes the local change in the function $f$ in terms of tangent vectors to the level set $f^{-1}(y_0)$ at the point $\bb{x}_0$.  In particular, since  $T_{\bb{x}_0}f^{-1}(y_0) \subset \mathbb{R}^n$ is a linear subspace of (the tangent space to) $\mathbb{R}^n$ and $\kdf(\bb{x}_0)$ has dimension $n-1$, it follows that the local dimension of $f$ is just one at almost every point in its domain.  Moreover, there is only one direction in $U$, the direction of the gradient $\nabla f(\bb{x}_0)$, in which $\bb{x}_0$ can move to produce any change in $f$ whatsoever.  This fact shows that the local structure of surjective $C^1$ functions is actually quite rigid, and there is the potential for constructing a (locally) one-dimensional representation of any such mapping despite the size of its domain.

The present work makes steps toward this idea by building on the ``Nonlinear Level set Learning'' (NLL) algorithm from \cite{NLL}, which is a neural network procedure for extrinsic dimension reduction in regression applications.  In particular, it is shown that the performance of this algorithm can be improved dramatically by incorporating the information provided by the IFT, which leads to less computational expense, faster training, and more accurate regression results.  To accomplish this, the NLL algorithm is recast as the minimization of a Dirichlet-type energy functional whose minimizers include mappings $\bb{g}:\mathbb{R}^n\to\mathbb{R}^n$, $\bb{g}(\bb{x}) = \begin{pmatrix} g_1(\bb{x}) & ... & g_n(\bb{x}) \end{pmatrix}$ which send the high-dimensional inputs to simple slices $\{\bb{g}(\bb{x})\,\,|\,\,g_1 = c \in \mathbb{R}\}$ where the function $f$ is constant, and which can be reliably approximated with neural network algorithms. Moreover, once a minimizer $\bb{g}$ has been computed, dimension reduction becomes a matter of simple truncation, and a one-dimensional regression can be performed to recover a model $\hat{f}:\mathbb{R}\to\mathbb{R}$ which predicts $f$ at any point in the original data space, i.e. $\hat{f}\circ g_1 \approx f$.  Experiments are provided which demonstrate the improved performance of this version of NLL over the original algorithm and over the state-of-the-art linear dimension reduction method Active Subspaces.

The remainder of the work is structured as follows:  Section 2 discusses related extrinsic dimension reduction methods and regression techniques, Section 3 details the NLL algorithm and its original formulation, Section 4 discusses the mentioned improvements to NLL and its connection to Dirichlet energy, and Section 5 exhibits numerical experiments which compare the present version of NLL to the original algorithm and Active Subspaces on a variety of test cases. Some concluding remarks are finally drawn in Section 6.

\section{Related Work}
It is well known that the regression problem \eqref{eq:reg} is difficult in the presence of sparse data, see e.g. \cite{hawkins2004, greenland2016} and references.  As such, there are a plethora of techniques which have been developed to mitigate the high response variability that is inevitable in this situation.  Ranging from optimal sampling (e.g. \cite{royall1970,dasgupta2009}) to 
central subspace methods (e.g. \cite{li2007,adragni2009}), all such works share the common themes of trying to increase predictive power and decrease model overfitting.  Extrinsic methods for dimension reduction in pointwise regression applications (e.g. \cite{ASbook, AM, NLL}) operate in the same way, although more regularity is usually assumed since a stronger type of convergence is discussed. 

The dimension reduction approach most similar to ours is known as approximation by ridge functions \cite{pinkus1997,chui1992,constantine2017}.  In particular, consider a linear projection $P:\mathbb{R}^n \to \mathbb{R}^k$ where $k\ll n$ and a (not necessarily differentiable) function $\hat{f}:\mathbb{R}^k\to\mathbb{R}$.  If the functions $\hat{f}, P$ satisfy
\[ f(\bb{x}) = \hat{f}(P\bb{x}), \]
for all $\bb{x}\in U$, then $f$ is called a (generalized) ridge function (c.f. \cite{constantine2017}).  Clearly, this definition implies that $f$ is constant on the kernel of $P$.

Using ridge functions for dimension reduction typically involves optimizing over the functions $\hat{f}, P$.  In particular, suppose the projection $P$ is given and consider computing a $\hat{f}$ which satisfies
\begin{equation}\label{eq:lowdreg}
    \hat{f}(\bb{x}) \in  \argmin_{g:\mathbb{R}^k\to\mathbb{R}} \nn{f(\bb{x}) - g(P\bb{x})}^2_2.
\end{equation}
If $f$ is nearly constant on the kernel of $P$, then $\hat{f}$ will be a reasonable pointwise approximation to the original function.  Moreover, 
the minimization in \eqref{eq:lowdreg} is usually much more feasible than the one in \eqref{eq:reg} when the input data is sparse, as the projection $\bb{x}^s \mapsto P\bb{x}^s$ naturally increases its density.

Of course, to compute a useful solution to \eqref{eq:lowdreg} it is necessary to have a projection mapping which adequately captures the change in $f$.  This usually involves constructing a low-dimensional ``response surface'' containing information about the dependence of $f$ on its independent variables (see e.g. \cite{li2007,adragni2009,constantine2017,khuri2010} and their references), and frequently employs techniques including principal component analysis, projection pursuit regression, kriging, and others \cite{friedman1981,wold1987,abdi2010,myers2016,constantine2014,o2020,ma2013,cook2007,weng2017}.  One popular algorithm for response surface construction among $C^1$ functions is known as Active Subspaces (AS) \cite{ASbook}: a procedure for determining the affine subspace of $U\subset\mathbb{R}^n$ where $f$ changes the most on average.  In particular, AS computes a Monte Carlo approximation to the covariance matrix $\bb{C} \coloneqq \mathbb{E}[\nabla f (\nabla f)^T]$,
so that the eigenvalue decomposition $\bb{C} \eqqcolon \bb{W}\bm{\Lambda}\bb{W}^T$ gives a global linear transformation of the input coordinates in terms of how much they affect the value of $f$. In the case that only the first $k$ eigenvalues are significant, this yields a suitable low-dimensional projection $P:\mathbb{R}^n \to \mathbb{R}^k$ in terms of the first $k$ columns $\bb{W}_A$ of $\bb{W}$.  More precisely, decomposing $\bb{W} = \left[\bb{W}_A \,\, \bb{W}_I\right]$ in terms of its ``active'' and ``inactive'' components, it can be shown that 
\[ \nn{f(\bb{x}) - \hat{f}\left(\bb{W}_A^T\bb{x}\right)}_2^2 \leq C\nn{\bb{W}_I}_2^2, \]
where $\hat{f}$ is a suitably chosen function (see e.g. \cite[Theorem 2]{constantine2017}) and $C = C(\rho)$ is a Poincar{\'e} constant depending on the density $\rho$.  Due to this fact and others, AS projections are known to be quite useful for dimension reduction, regression, and sensitivity analysis (see e.g. \cite{ASbook,lukaczyk2014,diaz2018,lam2020}). 

On the other hand, it is frequently the case that the eigenvalues of the covariance matrix $\bb{C}$ decay quite slowly, making a linear AS reduction ineffective for low-dimensional regression (c.f. Section~\ref{sec:highdim}).  This has motivated another line of work into nonlinear methods for extrinsic dimension reduction.  One such idea was introduced as the Active Manifolds (AM) algorithm \cite{AM}, which takes advantage of the local decomposition of $f$ afforded by the IFT.  In particular, AM aims to construct an integral curve $t \mapsto \bb{x}(t) \subset \mathbb{R}^n$ of the (normalized) gradient field, i.e. a solution to 
\[ \dot{\bb{x}} = \frac{\nabla f(\bb{x})}{\nn{\nabla f(\bb{x})}}, \qquad \bb{x}(0) = \bb{x}_0. \]
Once this ``active manifold'' has been constructed, it is necessarily the case that the entire range of $f$ on the set of level sets intersecting $\bb{x}(t)$ is represented simply by the values $f(\bb{x}(t))$, i.e. for every $\bb{y}\in U$ such that $f(\bb{y}) = c$ and  $f^{-1}(c) \cap \bb{x}(t) \neq \varnothing$ there is a value $t_0$ such that $c = f(\bb{y}) = f(\bb{x}(t_0))$.  Therefore, to determine the value of $f$ at any suitable point in the input space it is sufficient to know the values of $f$ along the 1-D curve $\bb{x}(t)$ as well as a projection map $\pi:\mathbb{R}^n\to\mathbb{R}$, $\pi(\bb{y}) = \{t \,|\, f(\bb{y}) = f(\bb{x}(t))\}$.  This algorithm has the benefit of zero intrinsic error (since there is no averaging involved), but comes with the significant challenge of computing the necessary projection map.

However, recent work has also shown that sufficiently deep neural networks have the ability to reproduce arbitrary measurable functions (see e.g. \cite{hornik1989}), motivating another line of research into dimension reduction.  Most network-based intrinsic methods to date are based on constructing an  autoencoder-decoder network (see e.g. \cite{hinton2006,wang2016}) which learns projection and expansion functions that compose to yield an approximation to the identity mapping on the input space.   The extrinsic methods which employ neural networks are more various, including the DrLIM method in \cite{hadsell2006} which computes an invariant nonlinear function 
mapping the input data evenly to a low-dimensional space, or the method of DIPnets (see \cite{o2020}) which uses AS in conjunction with projected neural networks for increased generalizability.

\section{The NLL Algorithm}

In contrast to the work previously mentioned, the present approach is based on a neural network algorithm introduced in \cite{NLL} called ``Nonlinear Level set Learning'' (NLL).  At its core, NLL uses neural network techniques to extend the idea of AS to more general transformations of the input data.  In particular, consider computing a  diffeomorphism (differentiable bijection with differentiable inverse) $\bb{g}:\mathbb{R}^n \to \mathbb{R}^n$, $\bb{z} = \bb{g}(\bb{x})$ and $\bb{h}\circ\bb{g} = \bb{I}_{\mathbb{R}^n}$, which separates the domain of the push-forward function $f\circ \bb{h}$ into global pairs $\bb{z} = (\bb{z}_A, \bb{z}_I)$ of ``active'' and ``inactive'' coordinates.  If $\bb{g},\bb{h}$ can be constructed such that the sensitivity of $f\circ\bb{h}$ to the coordinates $\bb{z}_I$ is sufficiently low, then it is reasonable to conclude that for any inactive coordinate $z^i\in\bb{z}_I$ the domain of $f\circ\bb{h}$ can be restricted to $\mathrm{Span}\{z^i\}^\perp$ with negligible impact on the function value. Provided this condition is satisfied, regression can be applied to obtain a lower-dimensional mapping $\hat{f}:\mathbb{R}^{|A|} \to \mathbb{R}$ such that $f(\bb{x}) \approx \hat{f}(\bb{z}_A)$. More precisely, given the function $\bb{g}$ and writing $\bb{z}_A\eqqcolon\bb{g}_A(\bb{x})$ to denote its first $|A|$ components, this means computing a generalized ridge function
\begin{equation}\label{eq:nllreg}
    \hat{f}(\bb{g}_A(\bb{x})) \in \argmin_{\varphi:\mathbb{R}^{|A|}\to\mathbb{R}} \nn{f(\bb{x}) - \varphi(\bb{g}_A(\bb{x}))}_2^2,
\end{equation}
where $\bb{g}$ acts as the (nonlinear) projection operator.  Note that once $\bb{g}$ has been obtained, computing the required $\hat{f}$ in \eqref{eq:nllreg} is automatically a more feasible regression problem than \eqref{eq:reg}, since almost all of the variation in $f$ is concentrated in the lower dimensional image $\bb{z}_A$.  Most importantly, the necessary projection from $\bb{z}$ to $\bb{z}_A$ is simple and canonical: after truncating the domain of $\bb{h}$ by the span of the inactive variables $\bb{z}_I$, the active variables $\{\bb{z}_A\}$ parameterize the low-dimensional inputs by definition.

\begin{remark}
In the original NLL formulation \cite{NLL}, regression on $f\circ\bb{h}$ is still performed on the full set of inputs $\bb{g}(\bb{x})$ without dimension reduction.  On the other hand, we find that this is not necessary when the mappings $\bb{g},\bb{h}$ are well-trained.  Therefore, the algorithm in Section 3 does not require the use of additional inputs beyond $\bb{z}_A\coloneqq z_1$.
\end{remark}

Of course, to make use of this idea it is necessary to have an efficient way to compute the diffeomorphism $\bb{g}$.  The authors of \cite{NLL} have demonstrated that carefully designed neural networks are well suited to this task, and a minimization procedure can be employed to obtain a mapping which attempts to concentrate the sensitivity of $f$ in some predefined number of active directions. The particular network architecture and appropriate notion of loss which support this approach will now be discussed, as well as the present modifications which improve the overall efficacy of the method.

\subsection{Network Architecture}
In \cite{NLL} as well as presently, the constrained minimization for $\bb{g}$ and its inverse $\bb{h}$ is accomplished using a specialized RevNet architecture \cite{RevNet,gomez2017} based on the Verlet discretization of Hamiltonian systems.  RevNets are neural networks which are reversible by construction, yielding improved memory efficiency as intermediate layer activations do not have to be stored.  The primary benefit of using RevNets inside the NLL algorithm is their connection to Hamiltonian systems, whose solution curves are generated by local diffeomorphisms of the source domain.  Because of this, propagating the input data through a RevNet structure (with a sufficiently small step-size) will necessarily give a configuration which is diffeomorphic to the original, removing the need for an explicit constraint during the minimization of the loss functional.

More precisely, let $(\bb{u},\bb{v})$ be a channel-wise partition of the inputs, $\bm{\sigma}$ be an activation function, and let $\bb{K}_i$ resp. $\bb{b}_i$ denote operator resp. vector valued functions for $i=1,2$. It is shown in \cite{RevNet} that the dynamical system 
\begin{align} \label{hamilsys}
\begin{split}
    \dot{\bb{u}}(t) &=  \bb{K}_1^T(t)\bm{\sigma}\left(\bb{K}_1(t)\bb{v}(t) + \bb{b}_1(t)\right), \\
    \dot{\bb{v}}(t) &= -\bb{K}_2^T(t)\bm{\sigma}\left(\bb{K}_2(t)\bb{u}(t) + \bb{b}_2(t)\right),
\end{split}
\end{align}
is stable and well-posed, hence usable for forward propagation along a neural network.  Defining $\bb{x} \eqqcolon (\bb{u}_0,\bb{v}_0)$ and discretizing
\eqref{hamilsys} with layers $1\leq l \leq L$ then yields the system
\begin{align*}
\bb{u}_{l+1} &= \bb{u}_l + \tau\,\bb{K}_{l,1}^T\, \bm{\sigma}\left(\bb{K}_{l,1}\bb{v}_l + \bb{b}_{l,1}\right), \\
\bb{v}_{l+1} &= \bb{v}_l - \tau\,\bb{K}_{l,2}^T\,\bm{\sigma}\left(\bb{K}_{l,2}\bb{u}_{l+1} + \bb{b}_{l,2}\right),
\end{align*}
where $\tau \in \mathbb{R}$ is the discrete time step, and $\bb{K}_{l,i}$ resp. $\bb{b}_{l,i}$ are the weight matrices resp. biases at layer $l$.  It is easily checked that this mapping is invertible at each layer, therefore it is reasonable to define $\bb{z} \coloneqq (\bb{u}_L, \bb{v}_L)$.  In practice, the partition $(\bb{u}_0, \bb{v}_0)$ for $\bb{x}$ is chosen so that $\bb{u}_0$ contains the first $\lceil n/2\rceil$ variables $x^i$ and $\bb{v}_0$ contains the remainder.   Propagation forward and backward through this scheme then yields mappings $\bb{g},\bb{h}$ which satisfy the desired diffeomorphism condition by construction, allowing for an unconstrained minimization of the loss functional.  The next goal is to discuss the particular loss criterion which is used to update the $\bb{K}_{l,i},\bb{b}_{l,i}$ parameters.

\subsection{The Loss Functional from \cite{NLL}}

It remains to discuss the criterion by which the mappings $\bb{g}, \bb{h}$ are trained.  The key observation to the approach in \cite{NLL} is that if $z^i \in \bb{z}_I$ is an inactive coordinate, then the gradient vector field $\nabla f(\bb{x})$ is orthogonal to the derivative of $\bb{h}(\bb{z})$ with respect to $z^i$ at any point $\bb{x} = \bb{h}(\bb{z})$ in the input space.  Said differently, this means that the derivative vector $\bb{h}_i(\bb{z}) \coloneqq \bb{h}'(\bb{z})\bb{e}_i$ is tangent to the level set of $f$ at $\bb{x}$, hence lies in the kernel of $f'(\bb{x})$.  Enforcing this condition during neural network training leads the authors of \cite{NLL} to the minimization problem
\begin{equation*}
    \argmin_{\bb{h}\in \mathrm{Diff}(\mathbb{R}^n)} \hat{L}(\bb{h}),
\end{equation*}
which involves the (regularized) loss functional
\begin{align}\label{nllloss1}
\hat{L}(\bb{h}) = \hat{L}_1 + \lambda\,\hat{L}_2 = \sum_{s\in S} \sum_{i=1}^n \omega_i \IP{}{\bb{J}_i(\bb{z}^s)}{\nabla f(\bb{x}^s)}^2 + \lambda \sum_{s\in S}\left(\mathrm{det}\,\bb{J}(\bb{z}^s) - 1\right)^2.
\end{align}
Here $\bb{J} = \begin{pmatrix} \bb{J}_1 & ... & \bb{J}_n \end{pmatrix} = \begin{pmatrix} \bb{h}_1/\nn{\bb{h}_1} & ... & \bb{h}_n/\nn{\bb{h}_n} \end{pmatrix}$ is the column-normalized Jacobian matrix of the transformation, the $\omega_i\in[0,1]$ are user-defined weights influencing the strength of the constraint in each dimension, and $\lambda\in[0,\infty)$ is a user-defined weight influencing the strength of the regularization term.
Note that the choice of column-normalization in $\bb{J}$ is introduced in order to simplify the calculation of derivatives in the original implementation, which uses a finite difference scheme.  Moreover, the regularization arises for similarly practical considerations, providing extra rigidity which helps direct the minimization toward a reasonable solution.  On the other hand, these additional inclusions come at the cost of diminishing the geometric meaning of the procedure, which has a substantial effect on both the rate of training and the overall effectiveness of the algorithm (c.f. Section~\ref{sec:numerics}).

\begin{remark}\label{rem:loss}
The informed reader will notice that the publicly available implementation of the NLL algorithm in \cite{NLL} uses a slightly different loss functional than $\hat{L}$ defined in \eqref{nllloss1}.  In particular, the loss functional minimized there is
\begin{equation}\label{nllloss2}
 \tilde{L}(\bb{h}) \coloneqq \frac{\sqrt{\hat{L}_1}}{|S|} + \lambda\prod_{s\in S} \left(\mathrm{det}\,\bb{J}(\bb{z}^s) - 1\right). 
 \end{equation}
Because the practical behavior of the original algorithm seems to be strongly dependent on this choice, each of the comparisons to ``Old NLL'' in Section~\ref{sec:numerics} uses whichever loss functional, \eqref{nllloss1} or  \eqref{nllloss2}, yields the best performance.
\end{remark}

While NLL in its original form has shown promising performance in several cases (c.f. \cite[Figure 2]{NLL}), it requires the specification many parameters $\omega_1, ..., \omega_n, \lambda$ whose significance is not clear and whose influence is not easily estimated {\em a priori}.  Additionally, it is relatively slow-to-train and requires the use of an expensive regularization term in order to guarantee reasonable performance and stability on high-dimensional data.  The goal of what follows is to describe theoretically-justified modifications to this algorithm which successfully eliminate these issues while also improving the overall efficacy of the dimension reduction.  By reformulating the NLL procedure as the minimization of an appropriate energy functional, a discretization is found which achieves both faster training and more accurate results.

\section{A Modified NLL Algorithm}\label{sec:newnll}
The present algorithm augments NLL with the geometric knowledge afforded by the IFT.  Consider using the same RevNet architecture to construct a bijective mapping  $\bb{h}:\mathbb{R}^n\to\mathbb{R}^n$ such that the level sets of $f$ are parameterized by the images of the inactive variables $\bb{z}_I = \{z^2, ..., z^n\}$, i.e. for each regular value $y\in\mathbb{R}$ there is a $c\in\mathbb{R}$ such that $f^{-1}(y) = \{\bb{h}(\bb{z})\,|\, z^1 = c\}$.  Since $f$ is differentiable and scalar-valued, the IFT asserts that this can be done locally away from points where $\nabla f = \bb{0}$, but there is no guarantee that a global mapping exists unless $\nabla f \neq \bb{0}$ everywhere and each level set in the domain is diffeomorphic to an $(n-1)$-dimensional hyperplane.  On the other hand, as with AS it is always reasonable to ask for a function  $\bb{h}$ which parameterizes these sets ``the most on average''.  Supposing $\bb{h}$ does this sufficiently well, the problem of approximating $f$ on the full input space can be reduced to that of approximating $f\circ \bb{h}$ on the one-dimensional space spanned by the active variable $z^1$.  In particular, an inexpensive regression approximation $\hat{f}:\mathbb{R}\to\mathbb{R}$ can be trained (e.g. simple neural network or local/global least-squares) using the information that $f(\bb{x}) \approx \hat{f}(g_1(\bb{x}))$.

\begin{remark}
If $\bb{v}_1, ..., \bb{v}_n $ form a frame of vector fields on an open set $U\in\mathbb{R}^n$, local coordinates $x^1,..,x^n$ such that $\bb{v}_i = \partial/\partial x^i$ are called simulatenous flow-box coordinates on $U$ \cite{bryant}.  Such coordinates exist around any point $\bb{x}\in U$ provided the vectors $\bb{v}_i(\bb{x})$ are linearly independent and the Lie bracket identities $[\bb{v}_i, \bb{v}_j] = 0$ hold on $U$ for all $1\leq i,j \leq n$.  In accordance with this notion, the NLL mapping $\bb{h}$ can be understood as providing approximate, best-on-average flow-box coordinates for the level sets of $f$.  Of course, it is easy to check that $[\bb{h}_i(\bb{z}), \bb{h}_j(\bb{z})] = \bb{h}'(\bb{z})[\bb{e}_i, \bb{e}_j] = \bm{0}$ for all $\bb{z}\in\mathbb{R}^n$.
\end{remark}

\subsection{A New Loss Functional}
Computing the mapping $\bb{h}$ in this way requires a loss functional which reflects the meaning of the procedure.  In practice, it suffices to note that if $f\circ \bb{h}$ is insensitive to perturbations in $z^2,...,z^n$ at a point $\bb{z} = \bb{g}(\bb{x})$, then its derivative $(f\circ \bb{h})'(\bb{z})$ is identically zero on the span of the inactive basis vectors $\{\bb{e}_2,...,\bb{e}_n\}$.  Note that this condition is readily expressed coordinate-wise as
\[(f\circ\bb{h})'(\bb{z})\bb{e}_i = \IP{}{\nabla f(\bb{x})}{\bb{h}_i(\bb{z})} = 0 \quad \mathrm{for\,all}\,\,i\neq 1,\] yielding precisely the motivating statement for the original NLL algorithm: that the vectors $\bb{h}_2(\bb{z}),...,\bb{h}_n(\bb{z})$ are orthogonal to $\nabla f(\bb{x})$. Conversely, note that the mathematical meaning of $(f\circ\bb{h})'(\bb{z})\bb{e}_i$ as a rate of change is preserved only if there is no normalization of $\bb{h}_i(\bb{z})$.  Therefore, given a set of training samples $\{\bb{x}^s, \nabla f(\bb{x}^s)\}_{s\in S}$ in the original input space and a RevNet architecture as before, the goal of our modified NLL algorithm is to compute $\bb{h}:\mathbb{R}^n\to\mathbb{R}^n$ which minimizes the loss functional
\begin{equation}\label{nllloss3}
    L(\bb{h}) = \frac{1}{|S|}\sum_{s\in S} \nn{(f\circ\bb{h})'(\bb{z}^s)}^2_\perp,
\end{equation}
where $\nn{\cdot}_\perp^2$ denotes the (squared) norm on the subspace orthogonal to the active direction $\bb{e}_1$ at each point, i.e. the trace of the Euclidean inner product $\IP{}{\cdot}{\cdot}$ with respect to the basis $\{ \bb{e}_2,...,\bb{e}_n \}$. This encourages the algorithm to find a mapping $\bb{h}$ which reduces the composite function $f\circ\bb{h}$ to a function of one variable $z_1 = g_1(\bb{z})$.  Moreover, since $\bb{h}$ is naturally constrained to be a diffeomorphism and hence a proper map, it follows that the summand is quadratic and strongly coercive (on $\mathrm{Span}\{\bb{e}_1\}^\perp$) whenever $f$ is, making gradient descent based on $L$ a feasible strategy. Note the similarity to the method of AS: while AS seeks a linear subspace of the input data $\{\bb{x}^s\}$ where the average change in $f$ is maximized, the proposed algorithm seeks a linear subspace $\mathrm{span}\{\bb{e}_2,...,\bb{e}_n\}$ of the transformed data $\{\bb{z}^s\}$ where the average change in the push-forward $f\circ\bb{h}$ is minimized.  It follows that the complementary subspace $\mathrm{span}\{\bb{e}_1\}$ maximizes this change, and the original nonlinear submanifold which maximizes the average change in $f$ can be recovered through $\bb{x} = \bb{h}(\bb{z})$.

\begin{remark}
Observe the lack of regularization in \eqref{nllloss3}.  Experiments show (c.f. Section 5) that this criterion is sufficient to drive the descent to a minimum without additional penalty, suggesting the benefits of using un-normalized derivatives of the network mapping.
\end{remark}

\subsection{Connection to Energy Minimization}

From a continuous perspective, it is meaningful to note that the $L$ defined in \eqref{nllloss3} is (up to scale) a discretization of the Dirichlet-type energy functional 
\begin{equation}
\mathcal{L}(\bb{h}) = \int_V \nn{(f\circ\bb{h})'(\bb{z})}_\perp^2\,d\mu^n = \int_I \int_{Z_t} \nn{(f\circ\bb{h})'(\bb{z})}_\perp^2\,d\mu^{n-1}\,dt, 
\end{equation}
where $\bb{h}(V) = U$, $I = \pi_1(V)$ is a bounded interval containing the range of $z_1$ and $Z_t = \{\bb{z}\in V\,|\,z^1 = t\}$.  To examine the structure of potential minimizers, it is useful to compute the variational derivative of $\mathcal{L}$.  Recall that a variation of $\bb{h}:V\to U$ is a one-parameter family of mappings (also denoted $\bb{h})$ satisfying $\bb{h}(t) = \bb{h} + t\bm{\varphi}$ for all $t$ in a compactly supported interval $t\in (-\varepsilon, \varepsilon)$ and some compactly supported $\bm{\varphi}\in C^1(V;U)$.   The variational derivative of a functional $\mathcal{F}$ depending on $\bb{h}$ is then the first-order term in its Taylor expansion around $t=0$.  In particular, there is the notation
$\delta\mathcal{F}(\bb{h})\bm{\varphi} = \frac{d}{dt}\mathcal{F}(\bb{h} + t\bm{\varphi})\big|_{t=0}$, 
which denotes the variational derivative (or simply the variation) of $\mathcal{F}$ at the point $\bb{h}$ in the direction of $\bm{\varphi}$.  It is a fundamental fact in the calculus of variations that $\mathcal{F}$ is stationary if and only if  $\delta\mathcal{F}(\bb{h})\bm{\varphi} = 0$ for all $\bm{\varphi}$.

Specifically to the present case, note that the variation in $\bb{h}$ induces a variation in $f\circ\bb{h}$,
\[ (f\circ\bb{h})(t) = (f\circ\bb{h}) + t(f\circ\bm{\varphi}), \]
so that the integrand of $\mathcal{L}$ varies as \[\delta \nn{(f\circ\bb{h})'(\bb{z})}^2_{\perp} = 2\IP{\perp}{(f\circ\bb{h})'(\bb{z})}{(f\circ\bm{\varphi})'(\bb{z})}.\]
Moreover, since $\IP{\perp}{\cdot}{\cdot}$ is just the inner product induced from $\mathbb{R}^{n}$ on the slices $Z_t$ and the variation $\bm{\varphi}$ vanishes on the boundary of each slice, integration by parts implies the equality
\[ \int_{Z_t} \IP{\perp}{(f\circ\bb{h})'(\bb{z})}{(f\circ\bm{\varphi})'(\bb{z})} \,d\mu^{n-1} = -\int_{Z_t} (f\circ\bm{\varphi})(\bb{z})\Delta^\perp (f\circ\bb{h})(\bb{z})\,d\mu^{n-1}. \]
Putting this together with the fact that spatial and variational derivatives commute in this setting, the variation of $\mathcal{L}$ is expressed as 
\begin{equation}\label{varia}
\delta\mathcal{L}(\bb{h})\bm{\varphi} = \int_I\int_{Z_t}\delta\nn{(f\circ\bb{h})'(\bb{z})}^2_{\perp}\,d\mu^{n-1}\,dt = -2\int_I\int_{Z_t}(f\circ\bm{\varphi})(\bb{z})\Delta^\perp (f\circ\bb{h})(\bb{z})\,d\mu^{n-1}\,dt. 
\end{equation}
Since $f\circ\bm\varphi$ is an arbitrary $C^1$ variation, this quantity \eqref{varia} vanishes identically if and only if $\Delta^\perp(f\circ\bb{h}) = 0$ on $V$, i.e. if and only if $f\circ\bb{h}$ is harmonic on the slices $Z_t$.  It is clear that this condition is satisfied when $\bb{h}$ parameterizes the level sets of $f$ on $U$, since $f\circ\bb{h}$ is constant on each slice.  Moreover, as minimization of Dirichlet-type energies such as $\mathcal{L}$ is known to exhibit good stability and convergence properties (see e.g. \cite{schoen1983}), it is reasonable to expect that the minimization of \eqref{nllloss3} will be similarly well-behaved.  The next Section provides experimental justification for this idea.

\section{Numerical Examples}\label{sec:numerics}
This Section details examples of the present NLL algorithm ``New NLL'' and its improvements over both AS and the original NLL algorithm ``Old NLL'' on sparse data sets. In particular, the effectiveness of each algorithm is measured using two metrics: {\em the relative sensitivity of the function to the active variable(s)}, and {\em the predictive ability of the low-dimensional approximation}.  After some discussion of implementation, two high-dimensional functions from \cite{NLL} are used for validation of New NLL.  Following this, New NLL is applied to the problem of predicting quantities of interest which depend on the solutions to systems of differential equations.  In all cases, New NLL is shown to effectively reduce the dimension to one, providing a low-dimensional submanifold which affords accurate approximation of the desired function.

\subsection{Implementation Details}
The relevant algorithms are implemented in Python on an early 2015 MacBook Pro with 2.7GHz Intel i5 processor and 8GB of RAM.  The implementation of AS is done in Python 2.7 following P. Constantine's code library \cite{AScode}, while the implementation of the NLL algorithms is done in Python 3.9 using the PyTorch library 1.7.  Note that the necessary derivatives of the network mapping $\bb{h}$ are computed using the PyTorch version of automatic differentiation, and not through finite differences as in \cite{NLL}.  Moreover, New NLL employs the Adaptive Moment Estimator (ADAM) optimizer during training \cite{Adam} while Old NLL uses stochastic gradient descent (SGD).  In both cases, the RevNet layer step size is fixed at $\tau=0.25$, the activation function is $\sigma = \tanh$, and 500 samples $\{\bb{x}, f(\bb{x}), \nabla f(\bb{x})\}$ are used for validation as the models train.  The amount of training samples is variable and reported in Tables~\ref{tab:toy} and \ref{tab:pde}.   The weights for Old NLL are chosen as in \cite{NLL}, namely $\lambda = 1$, $\omega_i = 1$ for $z^i$ an inactive variable and $\omega_i = 0$ otherwise. For visualization, the sensitivities of $f\circ\bb{h}$ (computed using gradient information) are reported as percentages relative to the sum over all coordinates, and the NLL  training and validation losses are reported as percentages relative to their initial values.

After dimension reduction, low-dimensional regression approximations are generated using the original training data projected onto the computed low-dimensional space.  To demonstrate that both traditional and modern regression methods can be used effectively after this dimension reduction, experiments on the functions $f_5, R_0$ below use local or global polynomial least-squares to approximate the function value while the experiments on $f_4, K$ use a simple feed-forward neural network. An additional 10000 uniformly distributed data samples are used for testing the regression in all cases except for $K$, where 500 additional samples are used. Results of the low-dimensional regressions are visualized through plots of the projected testing data against both the true and approximate function values.  The error metrics reported are relative root-mean-square error (RRMSE), relative $\ell_1$ error ($R\ell_1$), and relative $\ell_2$ error ($R\ell_2$), computed as
\[
    RRMSE(\hat{f}) = \frac{1}{\sqrt{|S|}}\left(\frac{\nn{f -\hat{f}}_2}{\max{f} - \min{f}}\right), \qquad R\ell_i(\hat{f}) = \frac{\nn{f-\hat{f}}_i}{\nn{f}_i}.
\]
Note finally that in the case of AS the mapping $\bb{h}$ should be interpreted as the matrix $\bb{W}^T$ from Section 2.  The results for all simulations in this Section are summarized in Tables~\ref{tab:toy} and \ref{tab:pde}.

\begin{table}[!htb]
\setlength\tabcolsep{1pt}
\resizebox{\columnwidth}{!}{%
\begin{tabular}{|l|l|llll|llll|llll|}
\hline
\multicolumn{2}{|l|}{} & \multicolumn{4}{l|}{100 Training Samples} & \multicolumn{4}{l|}{500 Training Samples} & \multicolumn{4}{l|}{2500 Training Samples} \\ \hline
Function & Method & \multicolumn{1}{l|}{$z_A$ Sens \%} & \multicolumn{1}{l|}{RRMSE \%} & \multicolumn{1}{l|}{R$\ell_1$ \%} & R$\ell_2$ \% & \multicolumn{1}{l|}{$z_A$ Sens \%} & \multicolumn{1}{l|}{RRMSE \%} & \multicolumn{1}{l|}{R$\ell_1$ \%} & R$\ell_2$ \% & \multicolumn{1}{l|}{$z_A$ Sens \%} & \multicolumn{1}{l|}{RRMSE \%} & \multicolumn{1}{l|}{R$\ell_1$ \%} & R$\ell_2$ \% \\ \hline
 & \cellcolor[HTML]{C0C0C0}New NLL & \cellcolor[HTML]{C0C0C0}78.7 & \cellcolor[HTML]{C0C0C0}3.86 & \cellcolor[HTML]{C0C0C0}8.27 & \cellcolor[HTML]{C0C0C0}10.9 & \cellcolor[HTML]{C0C0C0}89.8 & \cellcolor[HTML]{C0C0C0}1.82 & \cellcolor[HTML]{C0C0C0}3.52 & \cellcolor[HTML]{C0C0C0}5.16 & \cellcolor[HTML]{C0C0C0}94.5 & \cellcolor[HTML]{C0C0C0}0.827 & \cellcolor[HTML]{C0C0C0}1.72 & \cellcolor[HTML]{C0C0C0}2.35 \\
 & Old NLL & 60.4 & 6.63 & 14.5 & 18.8 & 65.9 & 4.58 & 10.5 & 13.0 & 69.2 & 4.02 & 9.11 & 11.4 \\
\multirow{-3}{*}{$f_4$} & AS 1-D & 25.8 & 30.3 & 75.9 & 85.9 & 25.9 & 21.7 & 39.5 & 61.4 & 25.9 & 15.9 & 37.6 & 44.8 \\ \hline
 & \cellcolor[HTML]{C0C0C0}New NLL & \cellcolor[HTML]{C0C0C0}75.1 & \cellcolor[HTML]{C0C0C0}0.920 & \cellcolor[HTML]{C0C0C0}5.79 & \cellcolor[HTML]{C0C0C0}7.92 & \cellcolor[HTML]{C0C0C0}88.6 & \cellcolor[HTML]{C0C0C0}0.370 & \cellcolor[HTML]{C0C0C0}2.78 & \cellcolor[HTML]{C0C0C0}3.97 & \cellcolor[HTML]{C0C0C0}93.8 & \cellcolor[HTML]{C0C0C0}0.154 & \cellcolor[HTML]{C0C0C0}1.63 & \cellcolor[HTML]{C0C0C0}1.98 \\
 & Old NLL 1 & 54.6 & 0.699 & 7.48 & 9.40 & 55.4 & 0.942 & 7.26 & 9.52 & 56.1 & 0.784 & 6.91 & 8.05 \\
\multirow{-3}{*}{$f_5$} & Old NLL 2 & 61.8 & 1.80 & 12.9 & 21.1 & 68.7 & 1.03 & 9.22 & 11.1 & 67.5 & 0.894 & 8.16 & 9.69 \\ \hline
\end{tabular}%
}
\caption{Results from the experiments in Section~\ref{sec:highdim}, including sensitivity measures and low-dimensional regression errors.}
\label{tab:toy}
\end{table}

\subsection{Two High-Dimensional Examples}\label{sec:highdim}
Consider the following functions from \cite{NLL}, numbered consistently with their source,
\begin{equation*}
    f_4(\bb{x}) = \sin\left(\nn{\bb{x}}^2\right), \qquad f_5(\bb{x}) = \prod_{i=1}^{20} \frac{1}{1 + x_i^2}.
\end{equation*}

\begin{remark}
Note that the $f_5$ in \cite{NLL} is actually $1.2^{40}f_5(1.2\bb{x})$ in terms of the present $f_5$, therefore identical up to a scale factor of $1.2^{40}$ and a dilation $\bb{x} \mapsto 1.2\bb{x}$.
\end{remark}

First, the experiment from \cite{NLL} on $f_5:[0,1]^{20} \to \mathbb{R}$ is repeated, which uses a 30-layer RevNet.  Here both Old NLL and New NLL are trained for 5000 epochs (passes forward and backward through the training data) with learning rates of 0.5 resp. 0.003, and regression is performed through local quadratic least-squares using the 10 nearest training neighbors of each test point. The results of New NLL are compared alongside simulations of Old NLL using one resp. two active variables, where Old NLL has been trained using the loss functional $\tilde{L}$ (c.f. Remark~\ref{rem:loss}).  

\begin{figure}[!htb]
% \captionsetup[sub]{justification=centering, labelfont=bf}
    \centering
    \begin{subfigure}{\textwidth}
    \centering
    \begin{minipage}{.04\textwidth}
    (a) 
    \end{minipage}
    \begin{minipage}{.95\textwidth}
    \includegraphics[width=\textwidth]{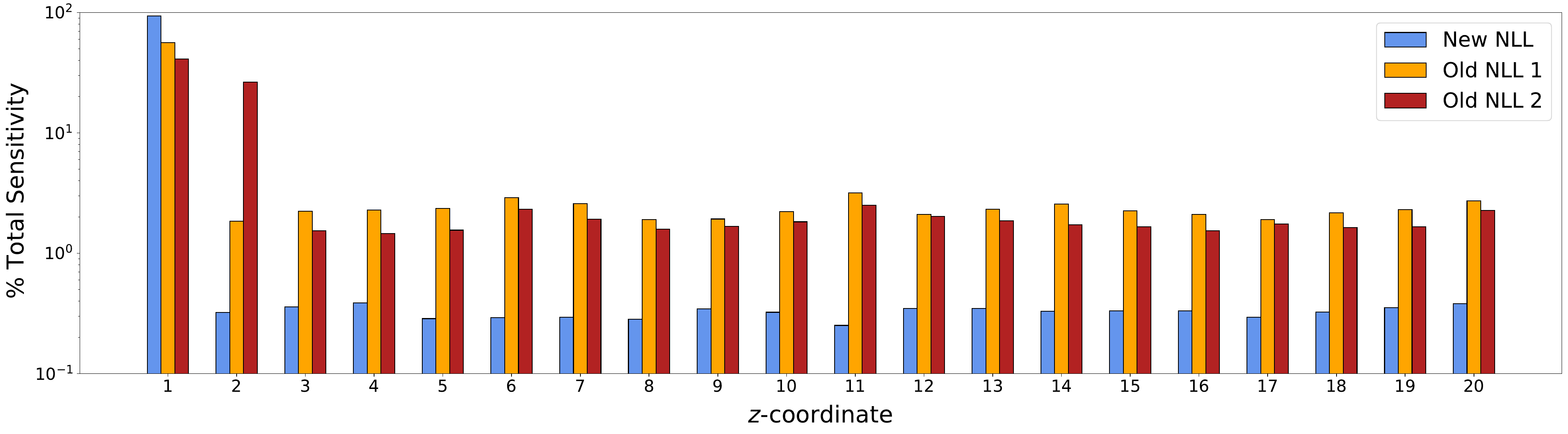} %was 0.68 textwidth
    \end{minipage}
    % \caption{Relative sensitivity of $f_5\circ\bb{h}$ to each $\bb{z}$-coordinate.}
    \end{subfigure}
    \begin{subfigure}{\textwidth}
    \centering
    \begin{minipage}{.04\textwidth}
    (b) 
    \end{minipage}
    \begin{minipage}{.95\textwidth}
    \includegraphics[width=\textwidth]{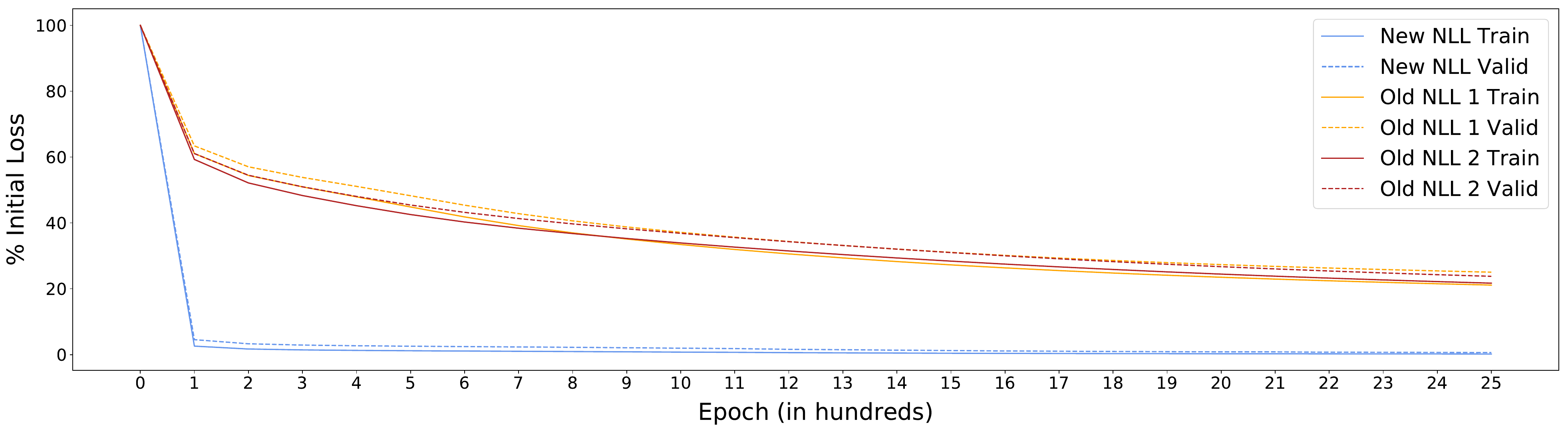} %was textwidth
    \end{minipage}
    % \caption{Relative value of loss during the first 2500 epochs of  NLL training.}
    \end{subfigure}
    \caption{(a) Relative sensitivity of $f_5\circ\bb{h}$ to each $\bb{z}$-coordinate;  
    (b) Relative value of loss during the first 2500 epochs of  NLL training.}
    \label{fig:ex5sensloss}
\end{figure}

\begin{figure}[!htb]
    \centering
    \begin{minipage}{.04\textwidth}
    (a) 
    \end{minipage}
    \,\,
    \begin{minipage}{.36\textwidth}
    \includegraphics[width=.8\textwidth]{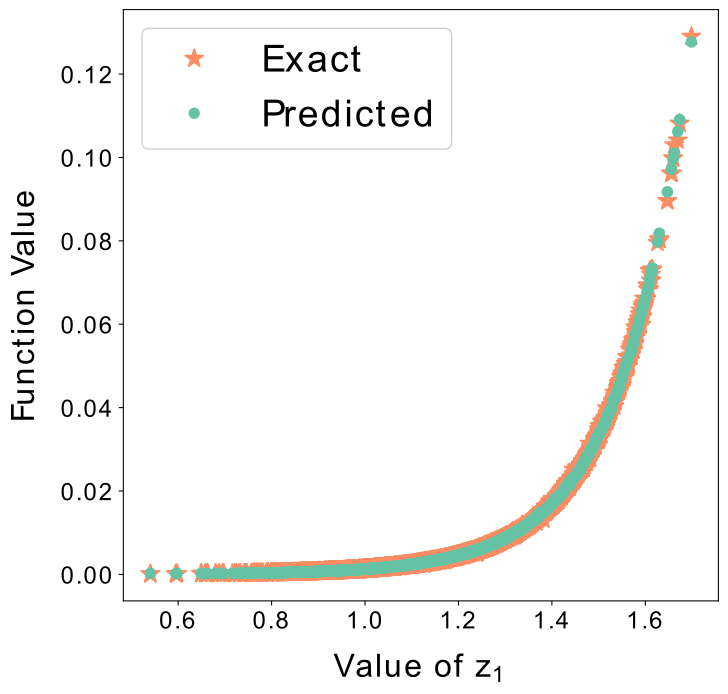}
    \end{minipage}
    %\hspace{1.5pc}
    \begin{minipage}{.04\textwidth}
    (b) 
    \end{minipage}
    \,\,
    \begin{minipage}{.36\textwidth}
    \includegraphics[width=.8\textwidth]{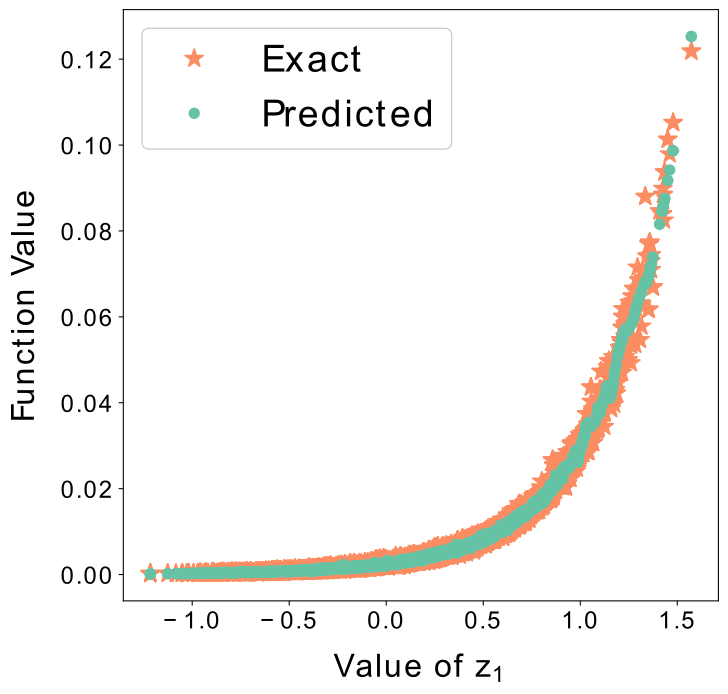}
	\end{minipage}
	
	\begin{minipage}{.04\textwidth}
    (c) 
    \end{minipage}
    \,\,
    \begin{minipage}{.36\textwidth}
    \includegraphics[width=0.80\textwidth, valign=t]{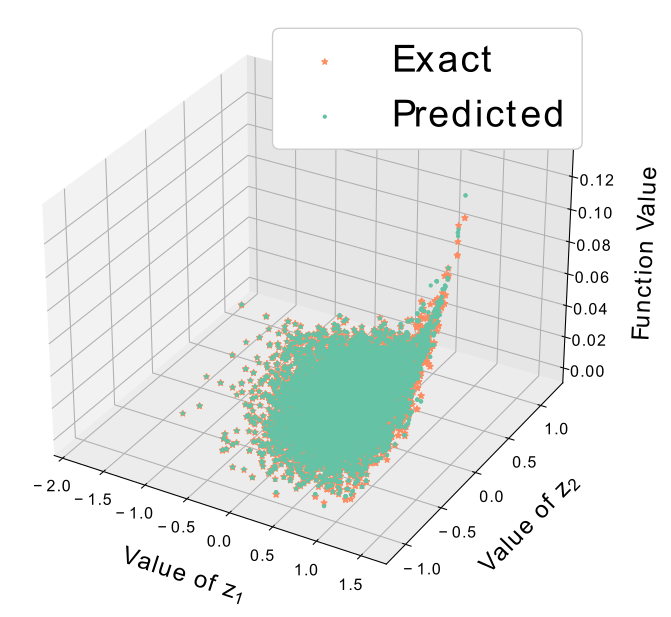}
    \end{minipage}
    \begin{minipage}{.04\textwidth}
    (d) 
    \end{minipage}
    \,\,
    \begin{minipage}{.36\textwidth}
    \includegraphics[width=0.80\textwidth, valign=t]{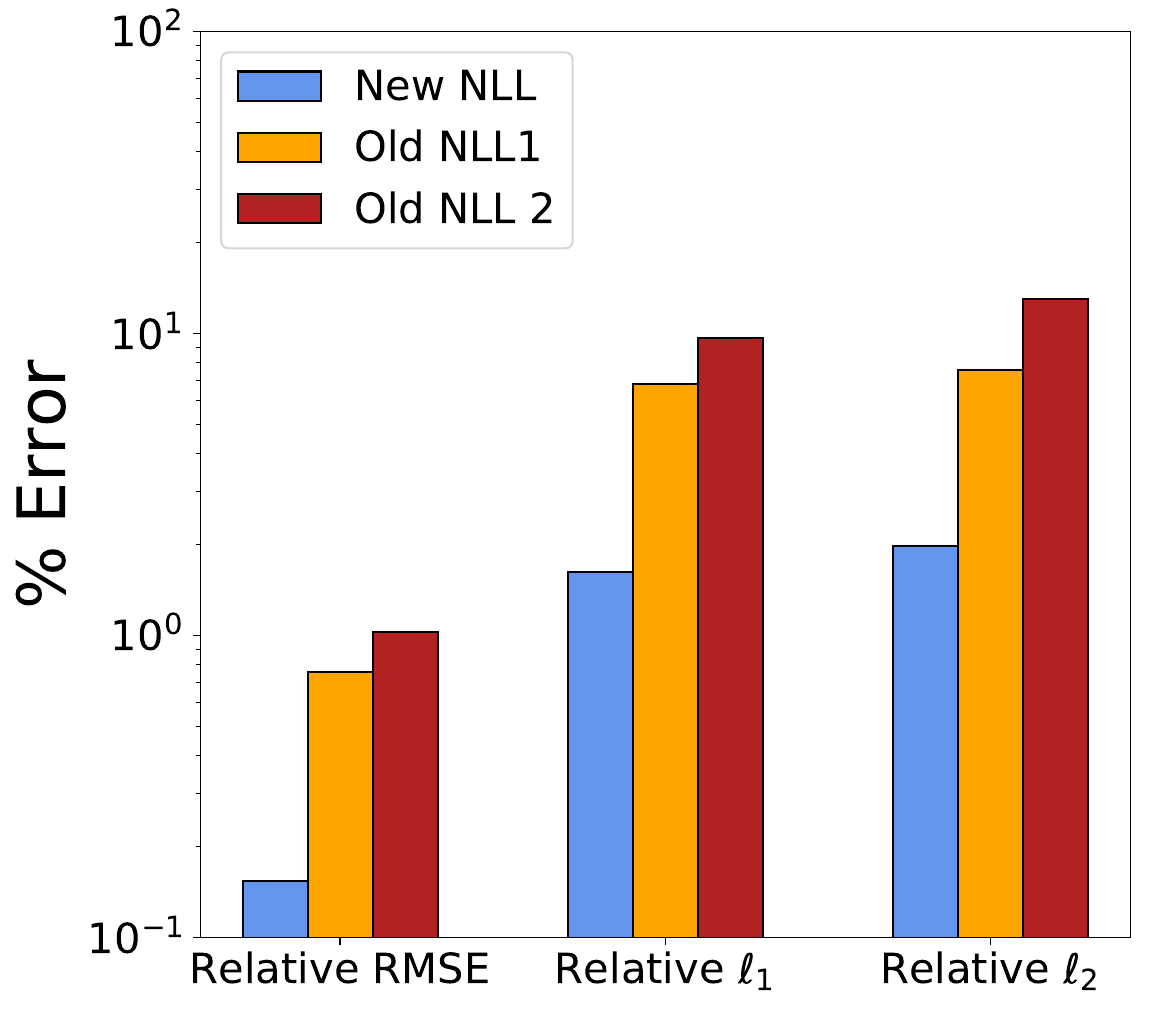}
    \end{minipage}
    \caption{Regression and errors on $f_5 \circ \bb{h}$: (a) New NLL; (b) Old NLL 1; (c) Old NLL 2; (d) Errors of the three approaches.
    }
    \label{fig:ex5reg}
\end{figure}

Figure~\ref{fig:ex5sensloss} shows that New NLL outperforms Old NLL in training speed (b) and sensitivity concentration (a).  In particular, with 500 training data New NLL concentrates 94\% of the total sensitivity in $f_5\circ\bb{h}$ in the $\bb{z}_1$ direction, while Old NLL can only achieve 56\%, or 67\% of the total if two active variables are used.  On the other hand, the regression results in Figure~\ref{fig:ex5reg} illustrate that low-dimensional approximations built using New NLL are also more accurate, as can be seen in both the regression errors as well as the tightness of the projected testing data around the fit curve.  Indeed, when using New NLL 500 training samples is sufficient for relative errors around 3\%. Interestingly, the inclusion of two active variables in Old NLL does not appear to improve the training rate nor the regression accuracy, despite concentrating more sensitivity in the active directions.

% \begin{figur

% \begin{figure}[h]
%     \centering
%     \begin{center}
%     \begin{subfigure}[b]{0.5\textwidth}
%     \includegraphics[width=\textwidth]{figs/f5sens.eps} %was textwidth
%     \caption{Relative sensitivity of $f_5\circ\bb{h}$ to each $\bb{z}$-coordinate.}
%     \end{subfigure}%
%     \begin{subfigure}[b]{0.5\textwidth}
%     \includegraphics[width=\textwidth]{figs/f5loss.eps} %was textwidth
%     \caption{Relative value of loss during first 2500 epochs of NLL training.}
%     \end{subfigure}
%     \end{center}
%     \begin{center}
%     \begin{subfigure}[b]{0.75\textwidth}
%     \begin{minipage}{0.33\textwidth}
%     \includegraphics[width=\textwidth]{figs/ex5_mine.png} %was textwidth
%     \end{minipage}%
%     \begin{minipage}{0.33\textwidth}
%     \includegraphics[width=\textwidth]{figs/ex5_theirs.png} %was textwidth
%     \end{minipage}%
%     \begin{minipage}{0.33\textwidth}
%     \includegraphics[width=\textwidth]{figs/ex5_theirs2d.png} %was textwidth
%     \end{minipage}%
%     \caption{Local least-squares regression: New NLL (left); Old NLL 1 (middle); Old NLL 2  (right).
%     }
%     \end{subfigure}%
%     \begin{subfigure}[b]{0.25\textwidth}
%     \includegraphics[width=\textwidth]{figs/f5Bars.eps} %was textwidth
%     \caption{Relative errors of the low-dimensional regression.}
%     \end{subfigure}
%     \end{center}
%     \caption{Results of experiment on $f_5$ with 2500 data.}
%     \label{fig:ex5}
% \end{figure}

Next, performance is measured on the higher-dimensional function $f_4:[0,1]^{40} \to \mathbb{R}$.  Again, a 30-layer RevNet is used, but now the regression is performed using a small feed-forward neural network of two fully-connected layers with 20 neurons each.  In each case, the regression is trained for 5000 epochs with a learning rate of 0.05, while the NLL networks Old NLL resp. New NLL are trained for 5000 epochs with learning rates of 0.02 resp. 0.003.  Again, the loss functional $\tilde{L}$ is used to train Old NLL.

\begin{figure}[!htb]
% \captionsetup[sub]{justification=centering, labelfont=bf}
    \centering
    \begin{subfigure}{\textwidth}
    \centering
    \begin{minipage}{.04\textwidth}
    (a) 
    \end{minipage}
    \begin{minipage}{.95\textwidth}
    \includegraphics[width=\textwidth]{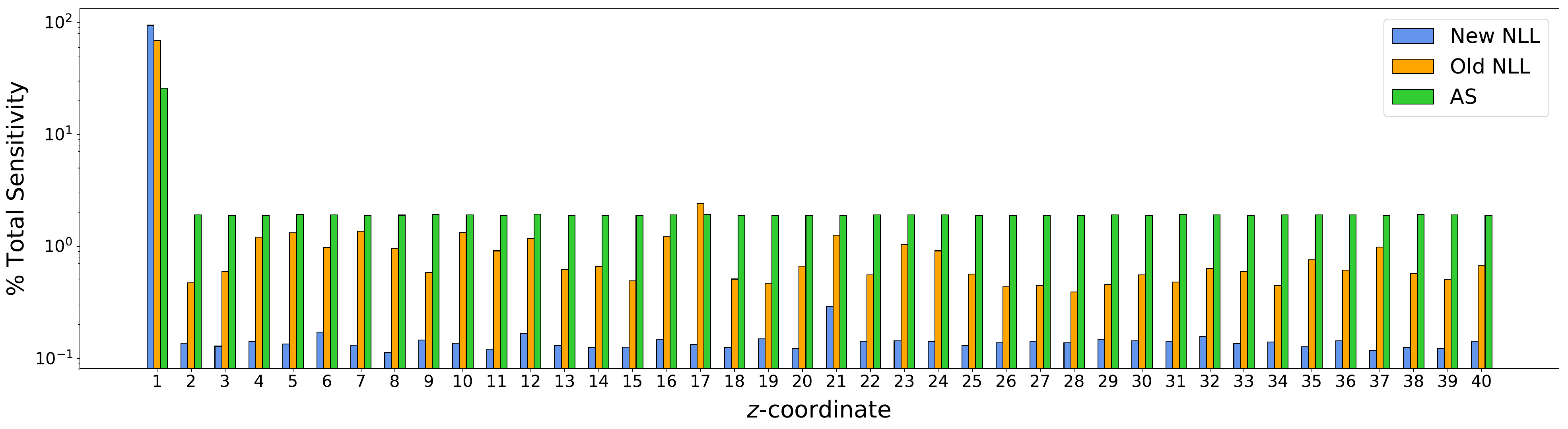} %was 0.68 textwidth
    \end{minipage}
    % \caption{Relative sensitivity of $f_4\circ\bb{h}$ to each $\bb{z}$-coordinate.}
    \end{subfigure}
    \begin{subfigure}{\textwidth}
    \centering
    \begin{minipage}{.04\textwidth}
    (b) 
    \end{minipage}
    \begin{minipage}{.95\textwidth}
    \includegraphics[width=\textwidth]{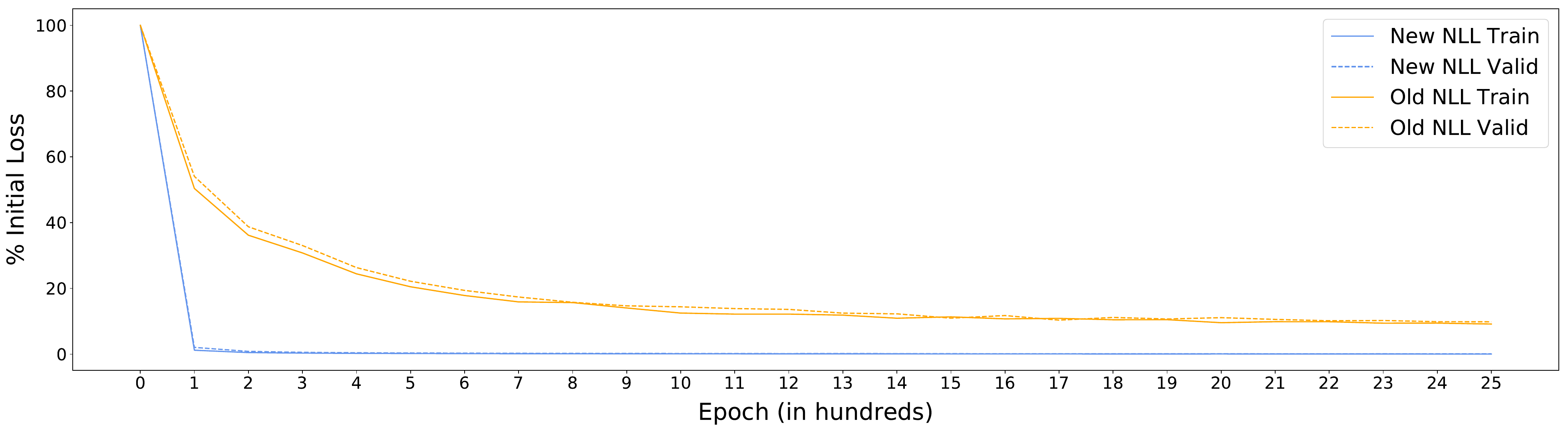} %was textwidth
    \end{minipage}
    % \caption{Relative value of loss during the first 2500 epochs of  NLL training.}
    \end{subfigure}
    \caption{(a) Relative sensitivity of $f_4\circ\bb{h}$ to each $\bb{z}$-coordinate; 
    (b) Relative value of loss during the first 2500 epochs of  NLL training.}
    \label{fig:ex4sensloss}
\end{figure}

\begin{figure}[!htb]
    \centering
    \begin{minipage}{.04\textwidth}
    (a) 
    \end{minipage}
    \,\,
    \begin{minipage}{.36\textwidth}
    \includegraphics[width=0.80\textwidth]{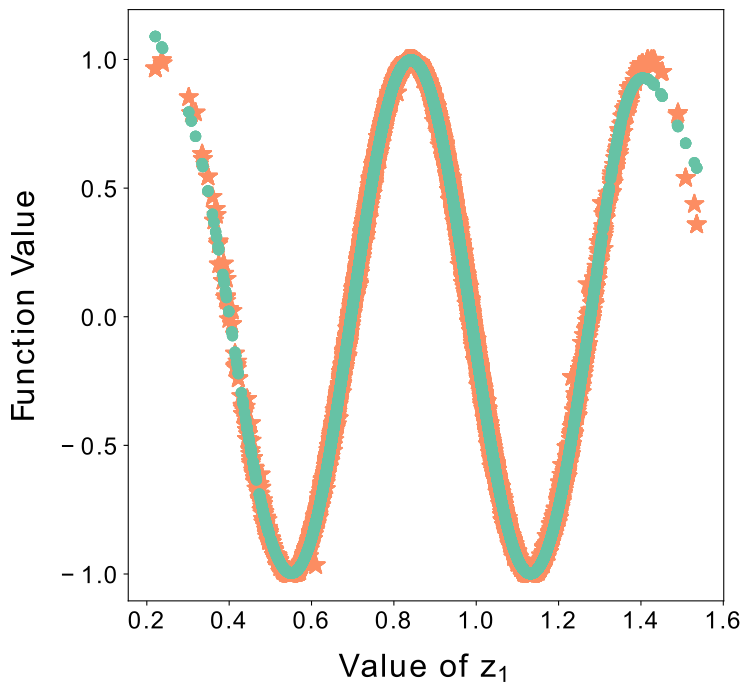}
    \end{minipage}
    \begin{minipage}{.04\textwidth}
    (b) 
    \end{minipage}
    \,\,
    \begin{minipage}{.36\textwidth}
    \includegraphics[width=0.80\textwidth]{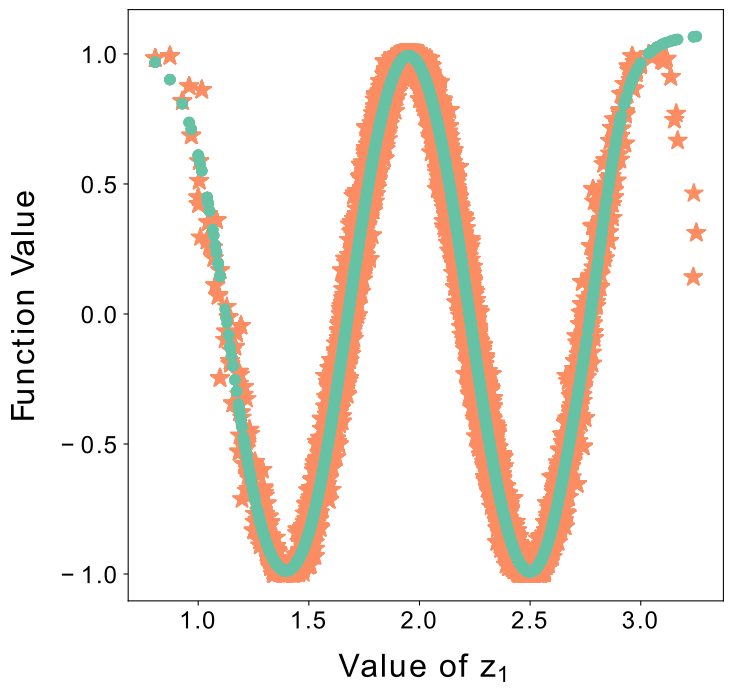}
    \end{minipage}

    \begin{minipage}{.04\textwidth}
    (c) 
    \end{minipage}
    \,\,
    \begin{minipage}{.36\textwidth}
    \includegraphics[width=0.80\textwidth, valign=t]{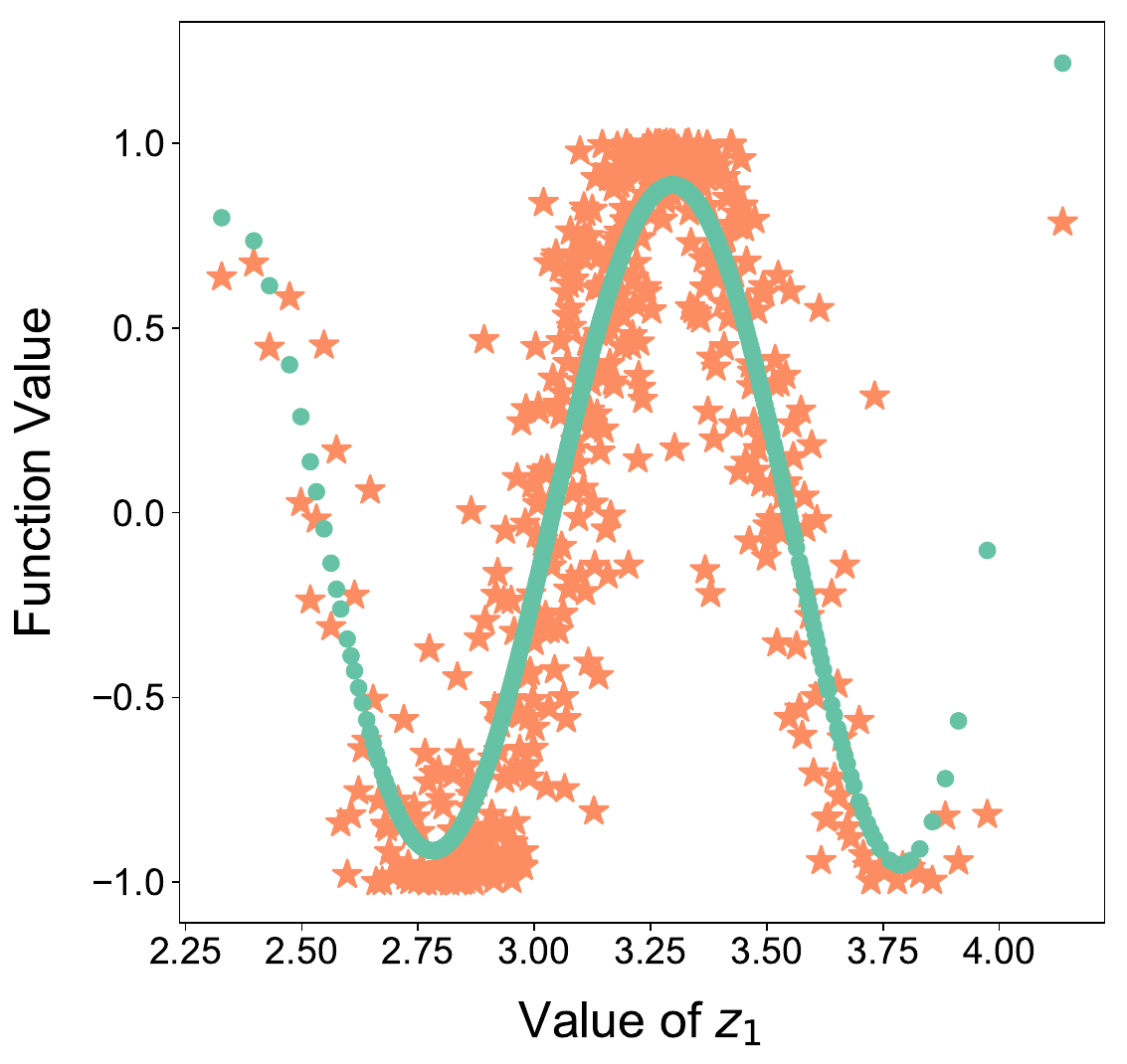}
    \end{minipage}
    \begin{minipage}{.04\textwidth}
    (d) 
    \end{minipage}
    \,\,
    \begin{minipage}{.36\textwidth}
    \includegraphics[width=0.80\textwidth, valign=t]{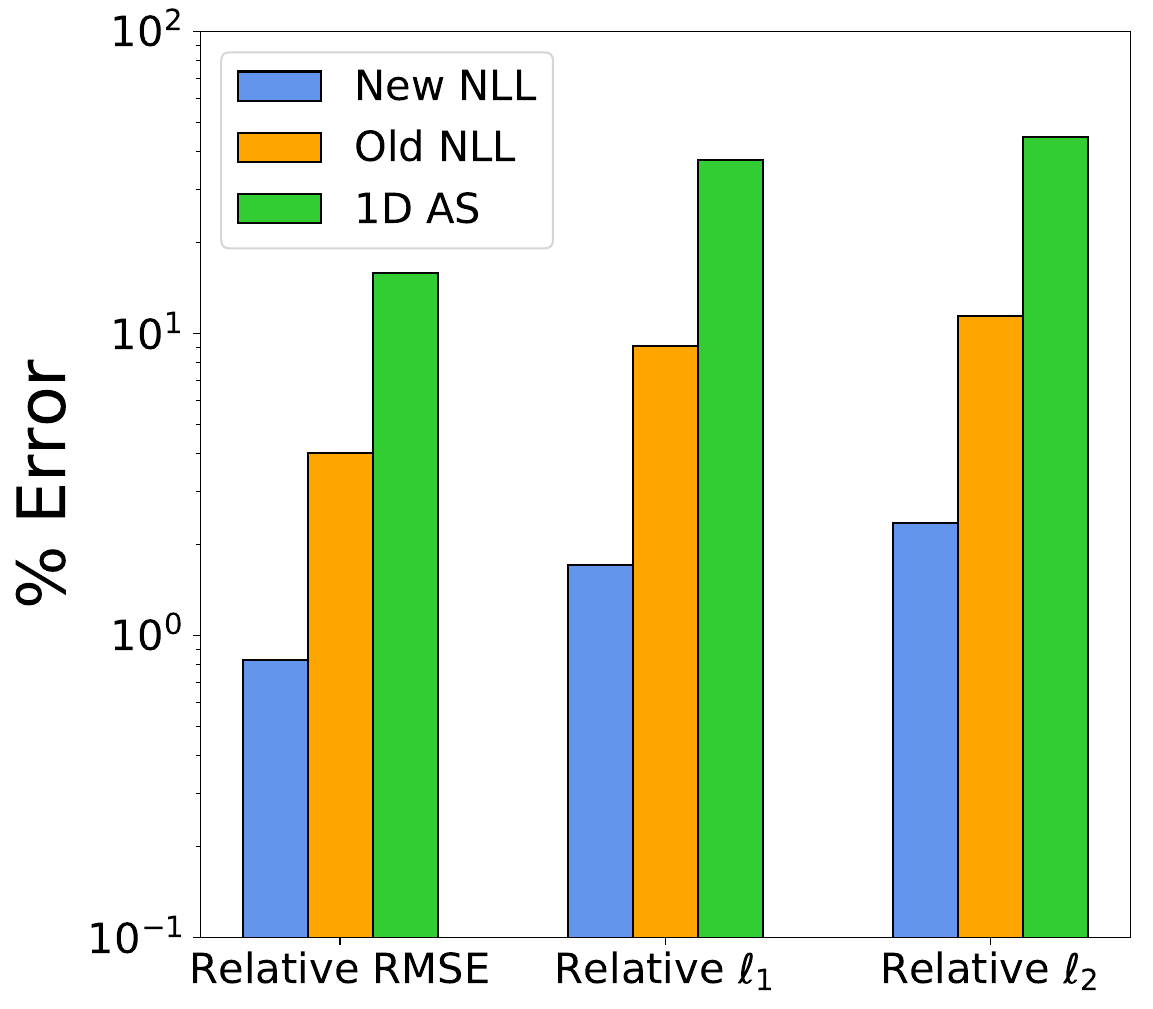}
    \end{minipage}
    \caption{Regression and errors on $f_4\circ\bb{h}$: (a) New NLL; (b) Old NLL; (c) Active Subspaces; (d) Errors of the three approaches.
    }\label{fig:ex4reg}
\end{figure}

The results of this experiment are illustrated in Figures~\ref{fig:ex4sensloss} and \ref{fig:ex4reg}.  Again, Figure~\ref{fig:ex4sensloss} (b) shows that New NLL reaches roughly 1\% of its initial training loss after just 100 epochs, while Old NLL plateaus around 15\% regardless of the training length.  The sensitivities of $f_4\circ\bb{h}$ also behave as expected, even in the presence of few training samples.  Indeed, for 500 samples New NLL concentrates 90\% of the total sensitivity in $\bb{z}_1$, Old NLL concentrates 66\%, and AS concentrates 26\%.  Note that the number of samples does not affect the quality of the AS reduction, which is both a strength and a weakness of this method.  Conversely, all algorithms benefit from increased training samples during the regression, although the approximation built using New NLL remains the most accurate.  Observe that 2500 samples is enough for around 2\% error with New NLL, while Old NLL still produces errors around 10\%.  Figure~\ref{fig:ex4reg} shows that Active Subspaces is also able to find the general shape of the function, but the one-dimensional approximation using this linear technique still produces around 40\% error.

% \begin{figure}[h]
%     \centering
%     \includegraphics[width=0.75\textwidth]{figs/f4sens.eps}
%     \begin{minipage}{0.59\textwidth}
%     \includegraphics[width=\textwidth]{figs/f4loss.eps}
%     \end{minipage}
%     \begin{minipage}{0.4\textwidth}
%     \includegraphics[width=\textwidth]{figs/f4Bars.eps}  
%     \end{minipage}
%     \begin{minipage}{0.3\textwidth}
%     \includegraphics[width=\textwidth]{figs/ex4_mine.eps}
%     \end{minipage}
%     \begin{minipage}{0.3\textwidth}
%     \includegraphics[width=0.98\textwidth]{figs/ex4_theirs.eps}
%     \end{minipage}
%     \begin{minipage}{0.3\textwidth}
%     \includegraphics[width=\textwidth]{figs/ex4_AS.eps}
%     \end{minipage}

\begin{table}[!htb]
\setlength\tabcolsep{1pt}
\resizebox{\columnwidth}{!}{%
\begin{tabular}{|l|l|llll|llll|llll|}
\hline
\multicolumn{2}{|l|}{} & \multicolumn{4}{l|}{20 Training Samples} & \multicolumn{4}{l|}{100 Training Samples} & \multicolumn{4}{l|}{500 Training Samples} \\ \hline
Function & Method & \multicolumn{1}{l|}{$z_A$ Sens \%} & \multicolumn{1}{l|}{RRMSE \%} & \multicolumn{1}{l|}{R$\ell_1$ \%} & R$\ell_2$ \% & \multicolumn{1}{l|}{$z_A$ Sens \%} & \multicolumn{1}{l|}{RRMSE \%} & \multicolumn{1}{l|}{R$\ell_1$ \%} & R$\ell_2$ \% & \multicolumn{1}{l|}{$z_A$ Sens \%} & \multicolumn{1}{l|}{RRMSE \%} & \multicolumn{1}{l|}{R$\ell_1$ \%} & R$\ell_2$ \% \\ \hline
 & \cellcolor[HTML]{C0C0C0}New NLL & \cellcolor[HTML]{C0C0C0}75.1 & \cellcolor[HTML]{C0C0C0}0.435 & \cellcolor[HTML]{C0C0C0}0.731 & \cellcolor[HTML]{C0C0C0}1.15 & \cellcolor[HTML]{C0C0C0}96.4 & \cellcolor[HTML]{C0C0C0}0.249 & \cellcolor[HTML]{C0C0C0}0.433 & \cellcolor[HTML]{C0C0C0}0.605 & \cellcolor[HTML]{C0C0C0}97.9 & \cellcolor[HTML]{C0C0C0}0.254 & \cellcolor[HTML]{C0C0C0}0.469 & \cellcolor[HTML]{C0C0C0}0.612 \\
 & Old NLL & 62.0 & 1.64 & 2.89 & 4.59 & 73.2 & 1.42 & 3.09 & 3.89 & 72.2 & 1.24 & 2.39 & 3.12 \\
 & AS 1-D & 55.0 & 9.18 & 18.5 & 22.1 & 54.4 & 18.2 & 22.0 & 9.14 & 53.6 & 18.8 & 22.7 & 9.43 \\
\multirow{-4}{*}{$R_0$} & AS 2-D & 79.3 & 4.34 & 7.87 & 10.4 & 79.7 & 8.01 & 10.8 & 4.49 & 79.6 & 8.26 & 10.8 & 4.50 \\ \hline
 & \cellcolor[HTML]{C0C0C0}New NLL & \cellcolor[HTML]{C0C0C0}{\color[HTML]{333333} 97.6} & \cellcolor[HTML]{C0C0C0}{\color[HTML]{333333} 0.425} & \cellcolor[HTML]{C0C0C0}{\color[HTML]{333333} 1.12} & \cellcolor[HTML]{C0C0C0}{\color[HTML]{333333} 1.27} & \cellcolor[HTML]{C0C0C0}98.3 & \cellcolor[HTML]{C0C0C0}0.186 & \cellcolor[HTML]{C0C0C0}0.496 & \cellcolor[HTML]{C0C0C0}0.555 & \cellcolor[HTML]{C0C0C0}{\color[HTML]{333333} 98.3} & \cellcolor[HTML]{C0C0C0}{\color[HTML]{333333} 0.101} & \cellcolor[HTML]{C0C0C0}{\color[HTML]{333333} 0.502} & \cellcolor[HTML]{C0C0C0}{\color[HTML]{333333} 0.540} \\
 & Old NLL & 80.1 & 3.52 & 9.82 & 10.5 & 80.5 & 3.19 & 8.99 & 9.53 & 80.3 & 3.25 & 9.15 & 9.70 \\
 & AS 1-D & 64.4 & 6.64 & 18.8 & 19.8 & 65.1 & 6.81 & 19.6 & 20.3 & 65.0 & 6.78 & 19.4 & 20.2 \\
\multirow{-4}{*}{$K$} & AS 2-D & 87.5 & 3.32 & 9.54 & 9.90 & 88.7 & 2.64 & 6.96 & 7.88 & 88.7 & 2.65 & 7.06 & 7.91 \\ \hline
\end{tabular}%
}
\caption{Results from the experiments in Section~\ref{sec:pde}, including sensitivity measures and low-dimensional regression errors.}
\label{tab:pde}
\end{table}

\subsection{Application to Parametric Differential Equations}\label{sec:pde}
As mentioned in Section 1, one of the primary applications of dimension reduction methods such as NLL and AS lies in predicting quantities of interest which arise from systems of differential equations.  To that end, we now consider two parameterized models for physical phenomena: a modified SIER model for disease spread, and an idealized model for fluid dynamics.

% \subsubsection{Predicting the Basic Reproduction Number of a Disease}

\paragraph{\bf Predicting the Basic Reproduction Number of a Disease.}
Let $\dot{f}$ denote the (total) time derivative of $f$ and consider the following modified SEIR model considered in \cite{diaz2018} for the spread of Ebola in West Africa

\begin{equation*}
\begin{aligned}[c]
\dot{S} &= -\beta_1SI - \beta_2SR_I - \beta_3SH,\\
\dot{E} &= \beta_1SI + \beta_2SR_I + \beta_3SH - \delta E,\\
\dot{I} &= \delta E - \gamma_1I - \psi I, \\
\dot{H} &= \psi I - \gamma_2H,
\end{aligned}
\qquad
\begin{aligned}[c]
\dot{R_I} &= \rho_1\gamma_1I - \omega R_I,\\
\dot{R_B} &= \omega R_I + \rho_2\gamma_2H,\\
\dot{R_R} &= (1-\rho_1)\gamma_1I + (1-\rho_2)\gamma_2H.
\end{aligned}
\end{equation*}

Here $S$ represents the fraction of the population that is susceptible to infection, $E$ represents the (infected but asymptomatic) exposed population, $I$ is the infected fraction, $H$ is the hospitalized fraction, $R_I$ represents the infectious dead (not properly buried), $R_B$ represents the non-infectious dead (properly buried), $R_R$ represents the recovered population, and $\delta = 1/9$ is a constant.  When studying the spread of disease, it is important to compute the basic reproduction number
\[R_0 = \frac{\beta_1 + \frac{\beta_2\rho_1\gamma_1}{\omega} + \frac{\beta_3}{\gamma_2}\psi}{\gamma_1 + \psi},\]
which depends on the eight parameters $\beta_1, \beta_2, \beta_3, \rho_1, \gamma_1, \gamma_2, \omega, \psi$ and measures the potential of the disease to transmit throughout the population. In \cite{diaz2018}, parameter ranges for this model are given around a baseline computed using data provided by the World Health Organization which was collected in the country of Liberia, and AS is used for the prediction of $R_0$. It is interesting to apply NLL to this problem for comparison with the existing results.  In this case, the training of the NLL models is done with a 15-layer RevNet and uniformly distributed samples drawn from the ranges in \cite[Table 7]{diaz2018}.  In particular, Old NLL resp. New NLL are trained for 5000 epochs with learning rates of 0.1 resp. 0.005, and regression is performed using 10-neighbor local quadratic least-squares.  Conversely, here global quartic least-squares are used for the AS regression, since global methods outperform local fitting when the spread of function values is wide (c.f. Figure~\ref{fig:ebolareg} (c)).   Note that Old NLL is trained using the loss functional $\hat{L}$, as this leads to better performance.

\begin{figure}[!htb]
% \begin{center}
\begin{center}
\begin{subfigure}[b]{0.5\textwidth}
	\begin{minipage}{.04\textwidth}
    (a) 
    \end{minipage}
    \,\,
    \begin{minipage}{.9\textwidth}
\includegraphics[width=\textwidth]{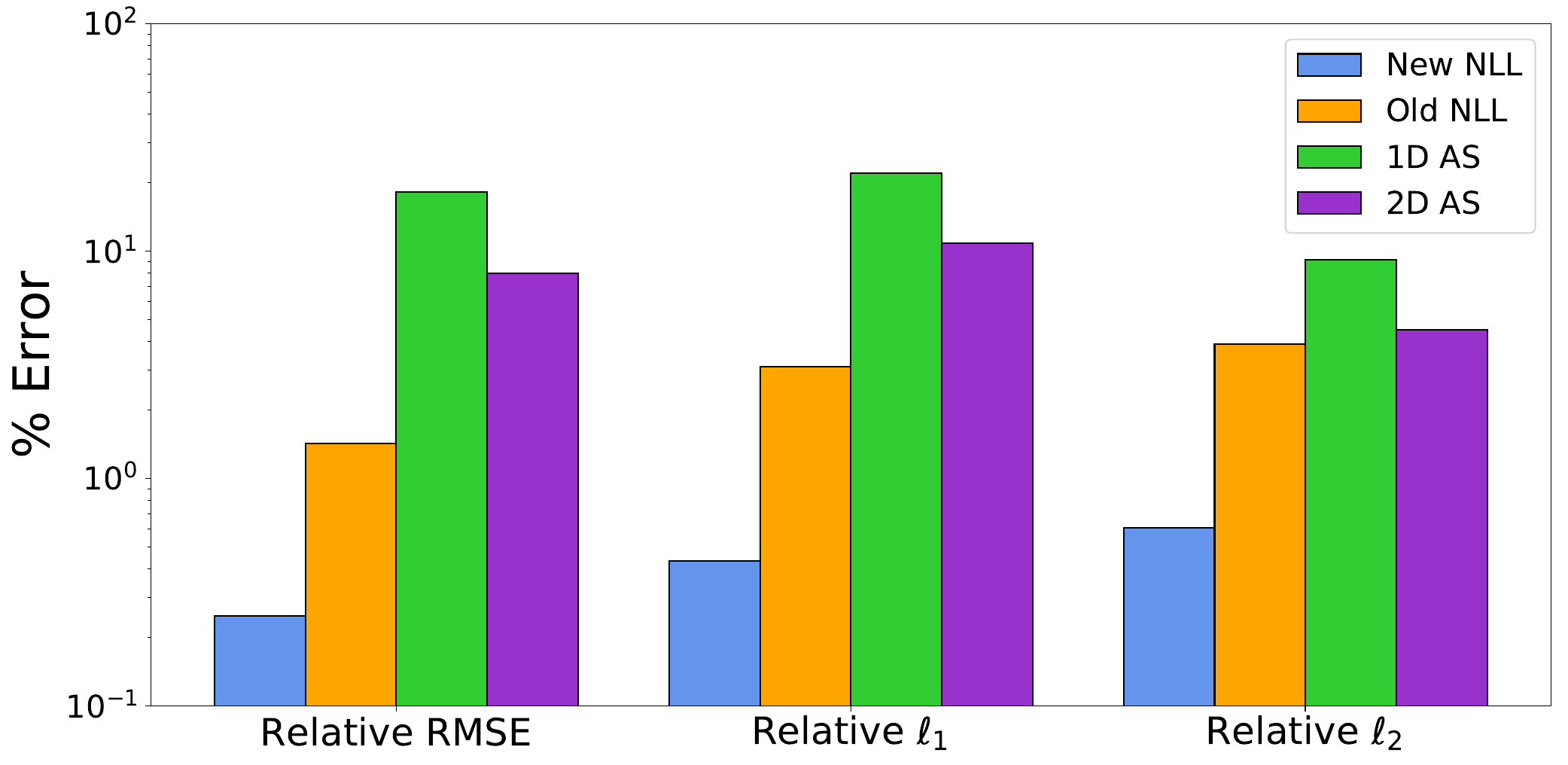}
	\end{minipage}
% \caption{Relative regression errors on $R_0\circ\bb{h}$.}
\end{subfigure}%
\begin{subfigure}[b]{0.5\textwidth}
% \begin{minipage}{\textwidth}
% \begin{center}
% \includegraphics[width=1\textwidth]{figs/ebolaPie.eps}
% \end{center}
% \end{minipage}
\,\,
	\begin{minipage}{.04\textwidth}
    (b) 
    \end{minipage}
    \begin{minipage}{.96\textwidth}
\includegraphics[width=1\textwidth]{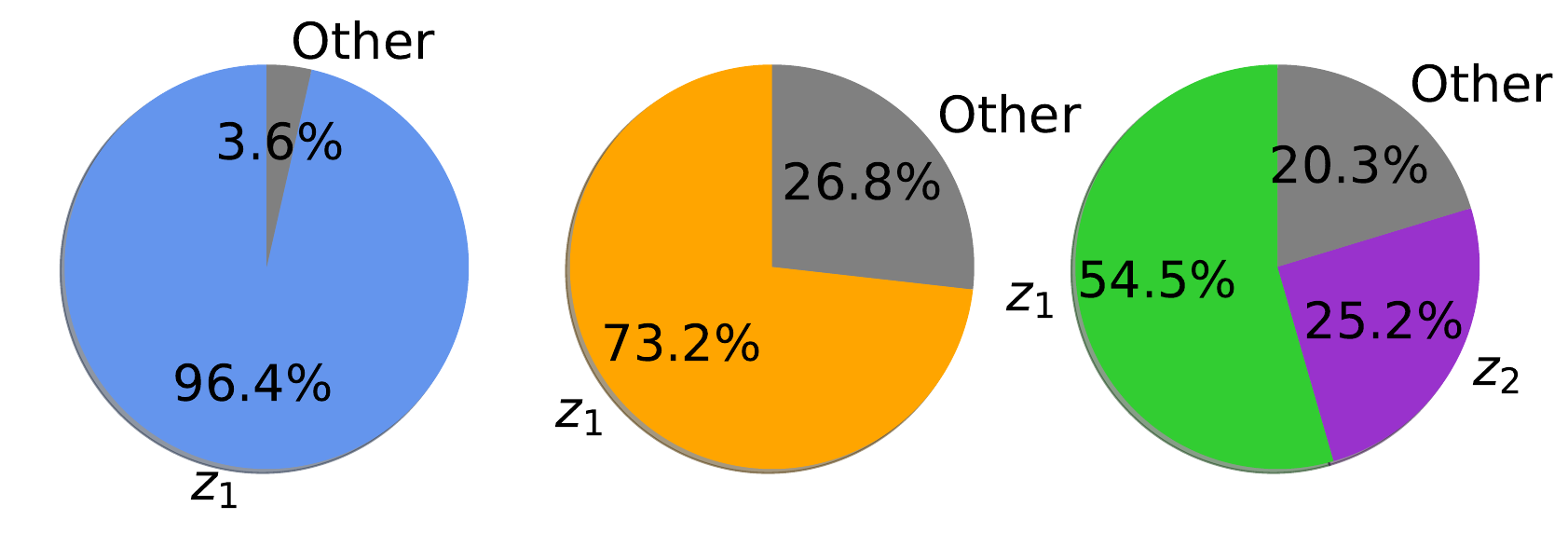}
	\end{minipage}
% \caption{Sensitivity of $R_0\circ\bb{h}$ to $z_1$ as a percentage of total.}
\end{subfigure}
\end{center}
\vspace{0.3pc}
\begin{subfigure}[b]{1\textwidth}
\centering
\begin{minipage}{.04\textwidth}
    (c) 
    \end{minipage}
    \begin{minipage}{.95\textwidth}
\includegraphics[width=\textwidth]{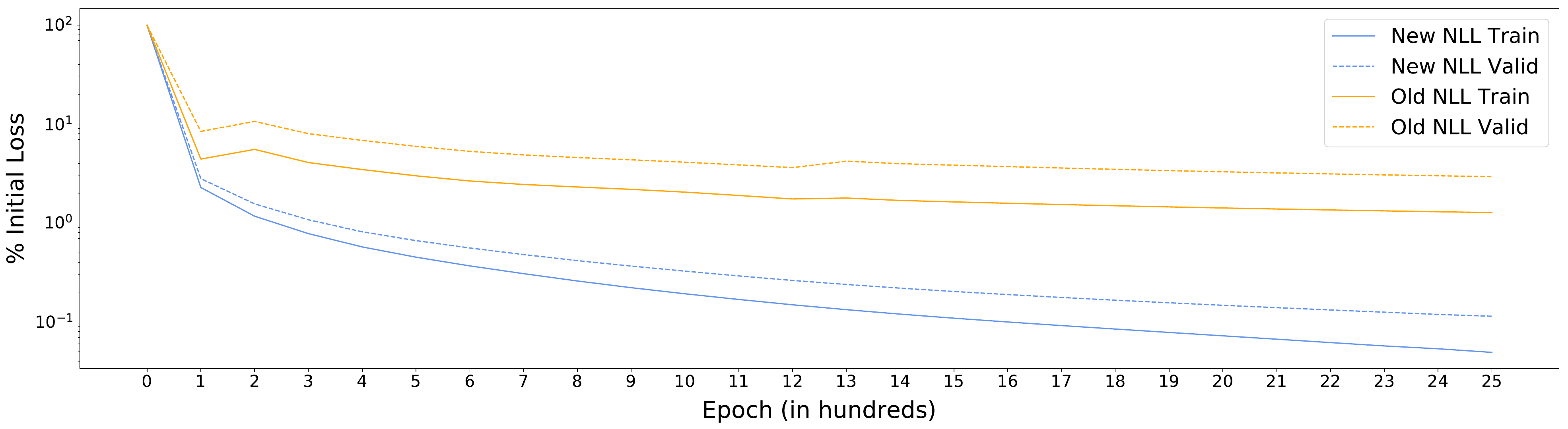}
	\end{minipage}
% \includegraphics[width=0.23\textwidth]{figs/ebola_mine.pdf}
% \includegraphics[width=0.23\textwidth]{figs/ebola_theirs.pdf}
% \includegraphics[width=0.23\textwidth]{figs/Ebola_1dAS.pdf}
% \includegraphics[width=0.25\textwidth]{figs/Ebola_2dAS.png}
% \caption{Relative value of loss during first 2500 epochs of NLL training (log scale).}
\end{subfigure}
% \end{center}
\caption{(a) Relative regression errors on $R_0\circ\bb{h}$; (b) Relative sensitivity of $R_0\circ\bb{h}$ to $\bb{z}_A$ as a percentage of total; (c) Relative value of loss during the first 2500 epochs of NLL training (log scale).}
\label{fig:ebolamore}
\end{figure}

\begin{figure}[!htb]
    \centering
    \begin{minipage}{.04\textwidth}
    (a) 
    \end{minipage}
    \,\,
    \begin{minipage}{.36\textwidth}
    \includegraphics[width=.8\linewidth]{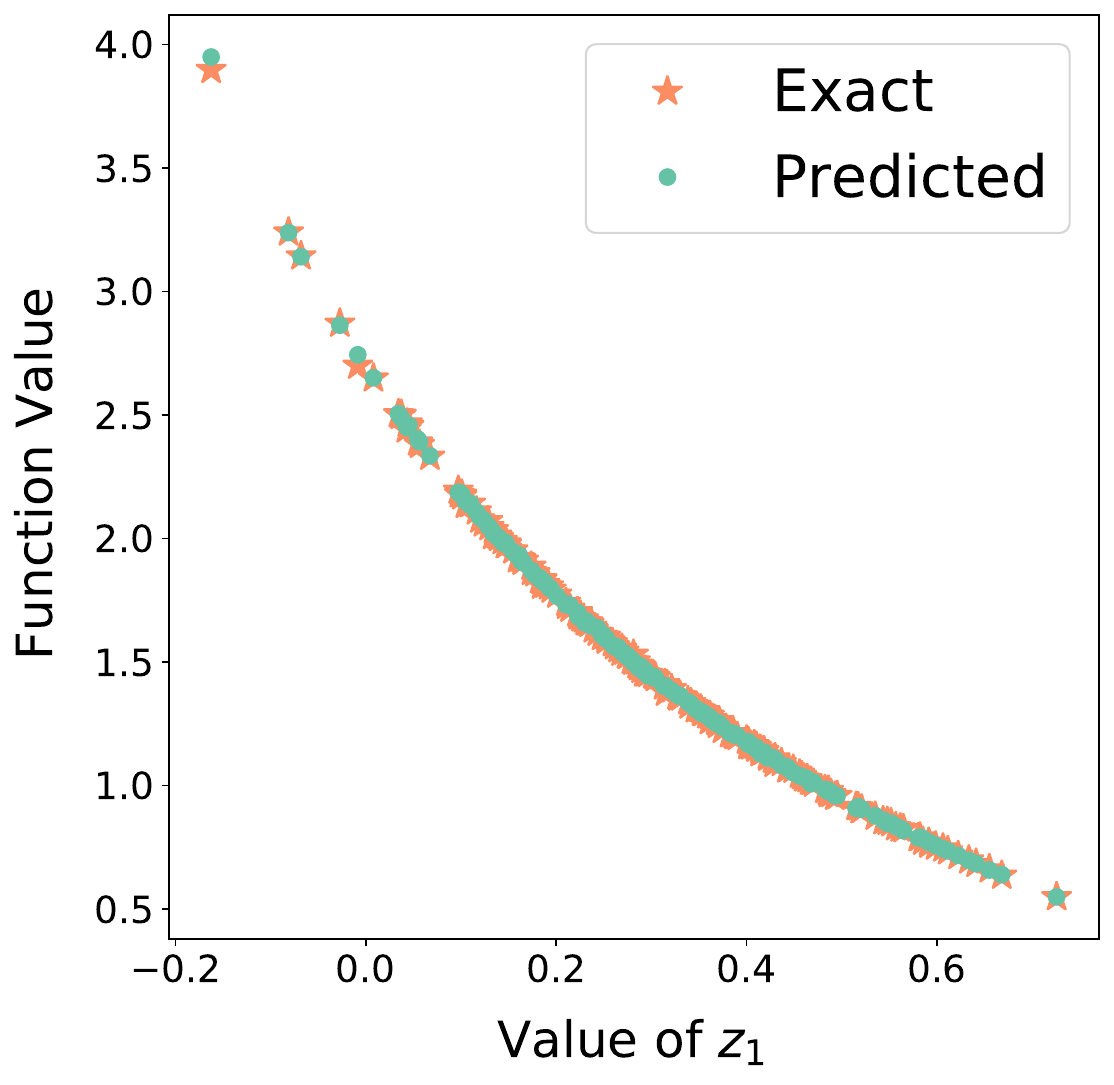}
    \end{minipage}
    \begin{minipage}{.04\textwidth}
    (b) 
    \end{minipage}
    \,\,
    \begin{minipage}{.36\textwidth}
    \includegraphics[width=.8\linewidth]{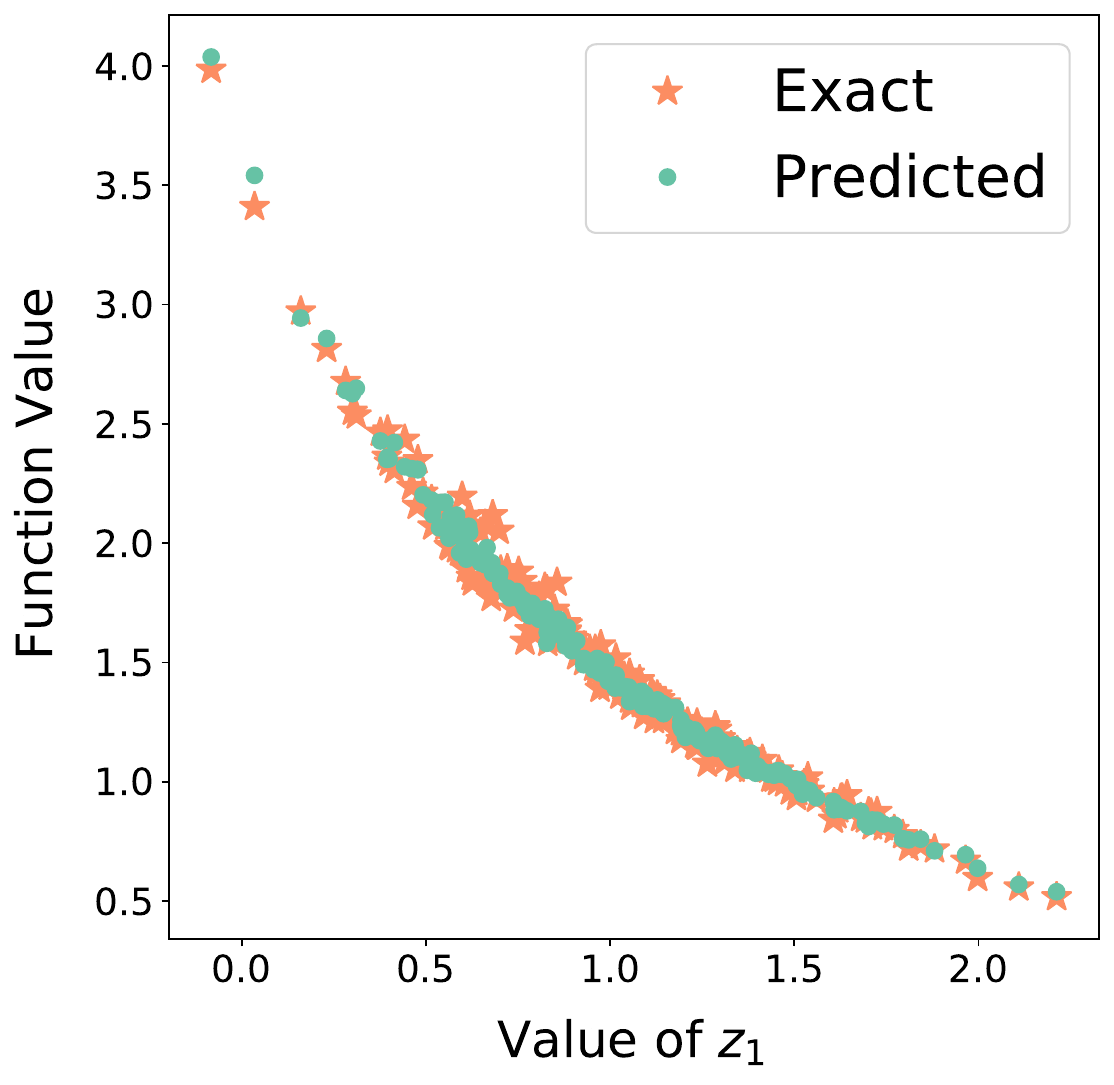}
    \end{minipage}
    
    \begin{minipage}{.04\textwidth}
    (c) 
    \end{minipage}
    \,\,
    \begin{minipage}{.36\textwidth}
    \includegraphics[width=.8\linewidth, valign=t]{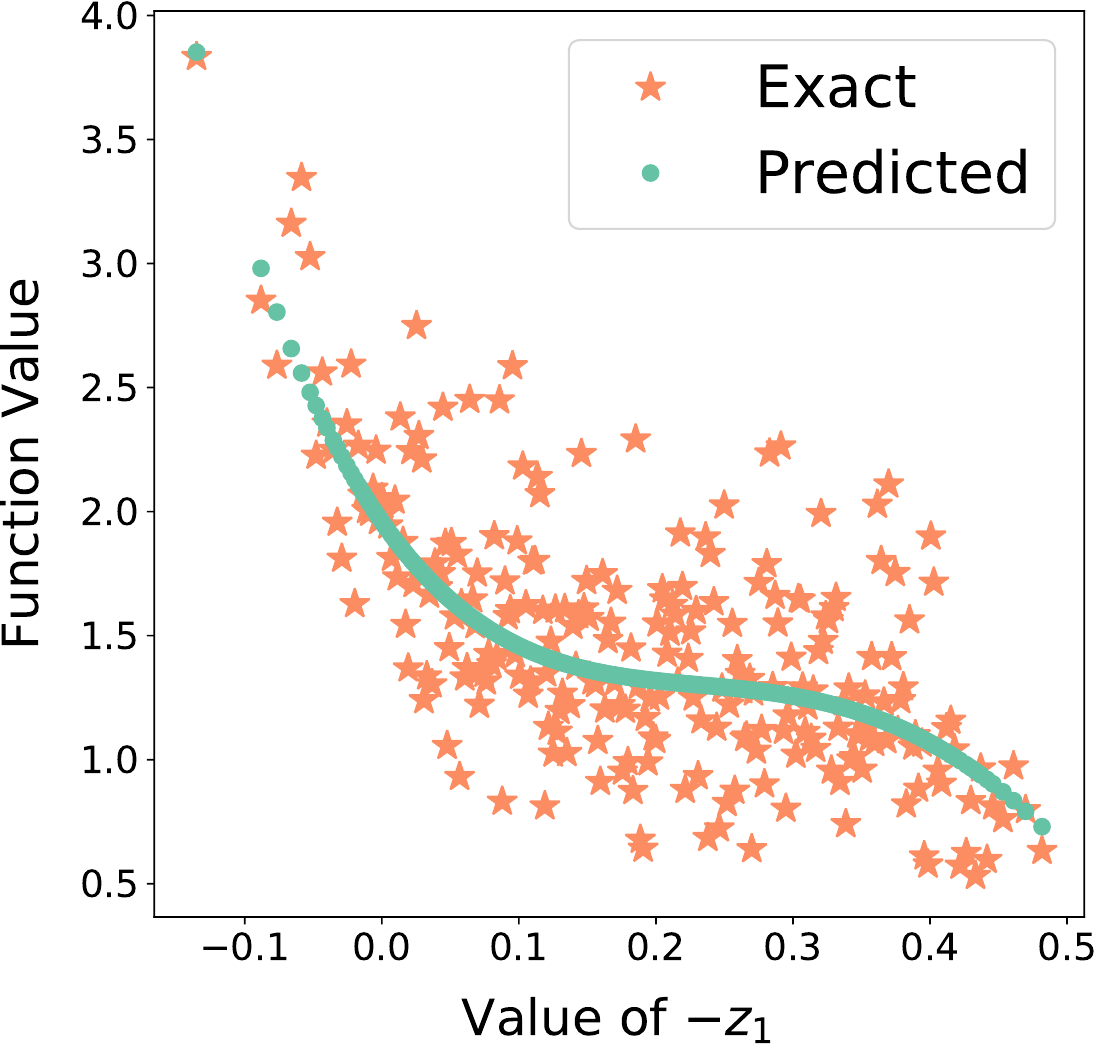}
    \end{minipage}
    \begin{minipage}{.04\textwidth}
    (d) 
    \end{minipage}
    \,\,
    \begin{minipage}{.36\textwidth}
    \includegraphics[width=.8\linewidth, valign=t]{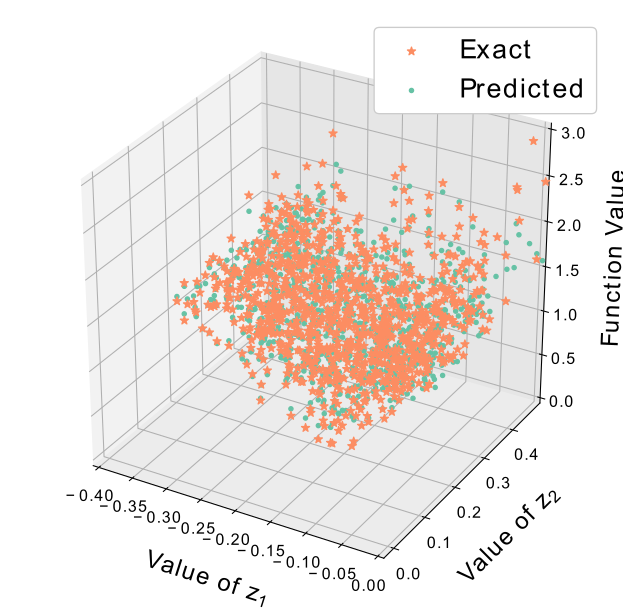}
    \end{minipage}
    \caption{Regression on $R_0\circ \bb{h}$: (a) New NLL; (b) Old NLL; (c) 1-D AS; (d) 2-D AS.}
    \label{fig:ebolareg}
\end{figure}

The results of this comparison show that the benefits of New NLL persist in this situation also.  In particular, 90\% of the total sensitivity in $R_0\circ \bb{h}$ is concentrated by New NLL in the active direction (for 100+ training samples), and even a 20 sample training set is sufficient for 1\% regression error.  Contrast this with Old NLL and either a one-dimensional or two-dimensional AS, which yield significantly more erroneous approximations.  Figure~\ref{fig:ebolareg} illustrates the low-dimensional regressions, and Figure~\ref{fig:ebolamore} (a) shows the errors.  As expected, Figure~\ref{fig:ebolareg} (a) shows that the approximation trained on the New NLL reduction produces a much tighter fit than existing methods.  Interestingly, note that here both NLL algorithms train relatively well (c.f. Figure~\ref{fig:ebolamore} (c)), and the Old NLL regression outperforms the 2-D AS regression despite concentrating less sensitivity in the active directions.

% In this case, the opposite of f5 happens in the sense that 2D AS produces worse regression despite concentrating more sensitivity.  It is noteworthy that both New and Old NLL train well ....   

% It is clear from the plots in Figure~\ref{fig:ebola} that the NLL algorithms produce a much more informative dimension reduction. Indeed, while the 2-D AS reduction is somewhat competitive with the 1-D Old NLL reduction, neither method reaches the performance of the New NLL reduction.

% \subsubsection{Predicting the Total Kinetic Energy}
\paragraph{\bf Predicting the Total Kinetic Energy.}
The last experiment in this Section considers the parameterized one-dimensional inviscid Burgers' equation, which is a common model for fluids whose motion can develop discontinuities.  In particular, let $\mmu = \left(\mu_1\,\, \mu_2\,\,\mu_3\right)^\intercal$ and consider the initial value problem,
\begin{align}\label{eq:burgers}
\begin{split}
    w_t + \frac{1}{2}\left(w^2\right)_x &= \mu_3 e^{\mu_2 x}, \\
    w(a, t, \mmu) &= \mu_1, \\
    w(x,0,\mmu) &= 1,
\end{split}
\end{align}
where $w=w(x,t, \mmu)$ represents the position of the fluid and $x\in[a,b]$.  It is interesting to examine the performance of NLL and AS in predicting the total kinetic energy at time $t$, given by 
\begin{align*}
    K(t, \mmu) &= \frac{1}{2}\int_0^t \int_a^b w(x,\tau,\mmu)^2 \,dx\,d\tau.
\end{align*}  
As $K$ is a function of the PDE solution, this represents a case where prediction is most valuable.  Indeed, for more expensive PDE simulations it may not be possible to generate as much data as desired.   Therefore, the goal is to use sparsely sampled values of $K$ obtained from simulations of \eqref{eq:burgers} to train an approximation to this function at any $(t, \mmu)$ in parameter space.

To accomplish this using the methods NLL and AS, it is necessary to have access to the gradient $\nabla K$ at the sampled points.  Noting that $[\partial_t, \partial_x] = [\partial_t, \partial_{\mu_i}] = [\partial_x, \partial_{\mu_i}] = 0$ for $1\leq i\leq 3$, simple differentiation yields
\begin{align*}
    \nabla K(t,\mmu) = \left( K_{t}\,\,K_{\mmu} \right)^\intercal &= \left( \frac{1}{2}\int_a^b w(x,t,\mmu)^2\,dx \quad \int_0^t\int_a^b w(x,\tau,\mmu)\,w_{\mmu}(x,\tau,\mmu)\,dx\,d\tau \right)^\intercal.
\end{align*}
It follows that $K_t$ can be obtained immediately from the solution $w$, and the components $K_{\bb\mu}$ are further computable by solving sensitivity equations.  In particular, differentiating (\ref{eq:burgers}) with respect to $\mmu$ yields the following system of initial value problems in the variable $w_{\mmu}$,
\begin{align}\label{eq:sensitivity}
\begin{split}
    w_{\mmu,t} + \left(ww_{\mmu}\right)_x &= \left( 0 \quad x\mu_3e^{\mu_2 x} \quad e^{\mu_2 x} \right)^\intercal \\
    w_{\mmu}(a, t, \mmu) &= (1\quad 0 \quad 0)^\intercal \\
    w_{\mmu}(x,0,\mmu) &= \bb{0}.
\end{split}
\end{align}
Solving (\ref{eq:burgers}) and (\ref{eq:sensitivity}) provides the necessary samples $\{K(t^s, \mmu^s),\, \nabla K(t^s, \mmu^s)\}_{s\in S}$ for applying the NLL and AS algorithms.

To study the example at hand, the parameters are chosen to take values in $(t, \mu_1, \mu_2, \mu_3) \in [25, 30] \times [3, 8] \times [0.015, 0.06] \times [0, 0.05]$ with $x\in [0,100]$.  Systems \eqref{eq:burgers} and  \eqref{eq:sensitivity} are discretized using simple forward Euler with upwinding with increments of $\Delta x= 0.4$ and $\Delta t = 0.025$.  Old NLL resp. New NLL are trained for 5000 epochs using a 7-layer RevNet with learning rates of 0.1 resp. 0.005, and regressions are performed using a neural network as in the case of function $f_4$.  Again, Old NLL is trained using the functional $\hat{L}$.

\begin{figure}[!htb]
\begin{center}
\begin{subfigure}[b]{0.5\textwidth}
	\begin{minipage}{.04\textwidth}
    (a) 
    \end{minipage}
    \,\,
    \begin{minipage}{.9\textwidth}
\includegraphics[width=\textwidth]{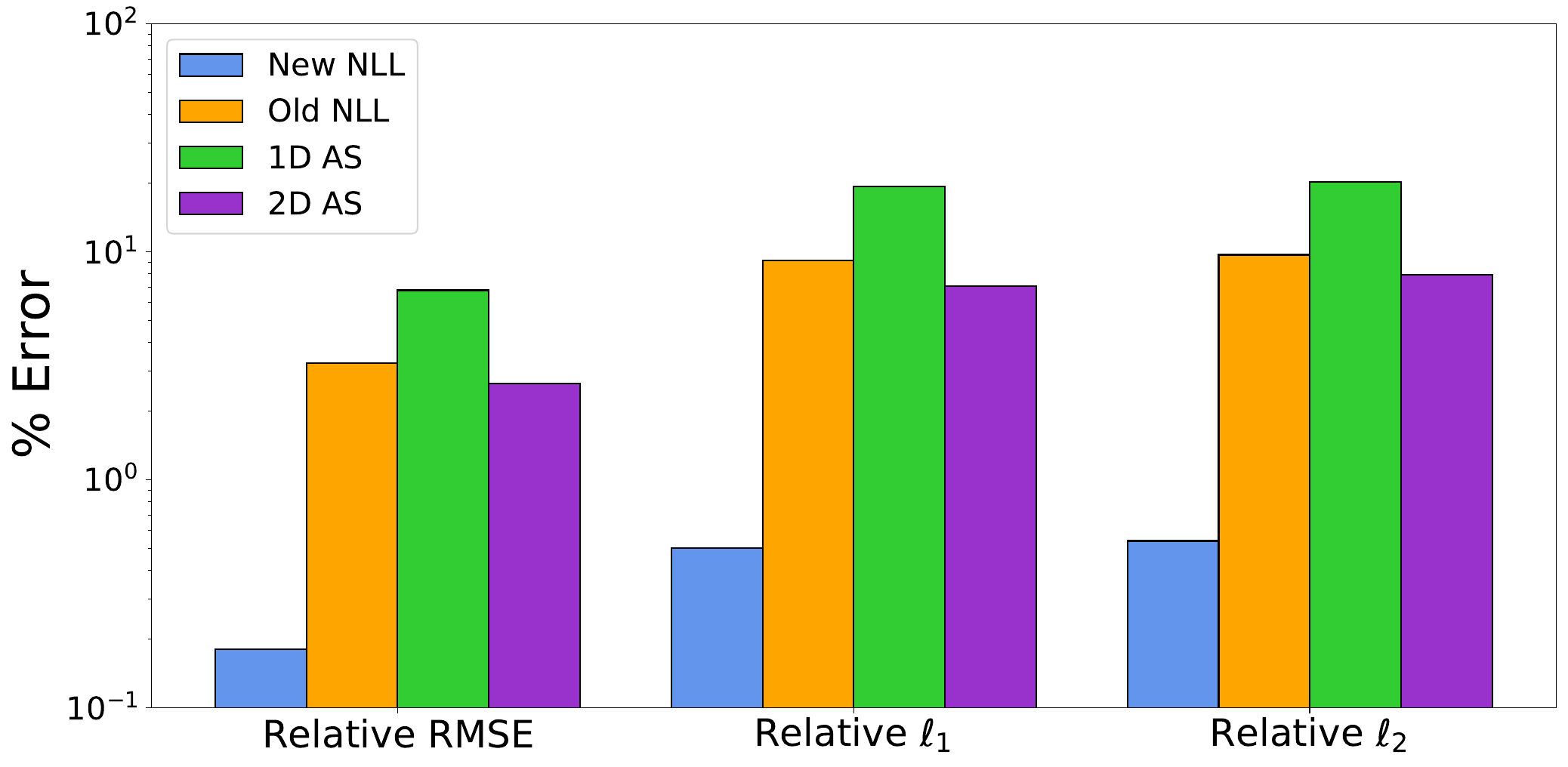}
	\end{minipage}
% \caption{Relative errors of the low-dimensional regression.}
\end{subfigure}%
\begin{subfigure}[b]{0.5\textwidth}
	\begin{minipage}{.04\textwidth}
    (b) 
    \end{minipage}
    \begin{minipage}{.96\textwidth}
\includegraphics[width=1\textwidth]{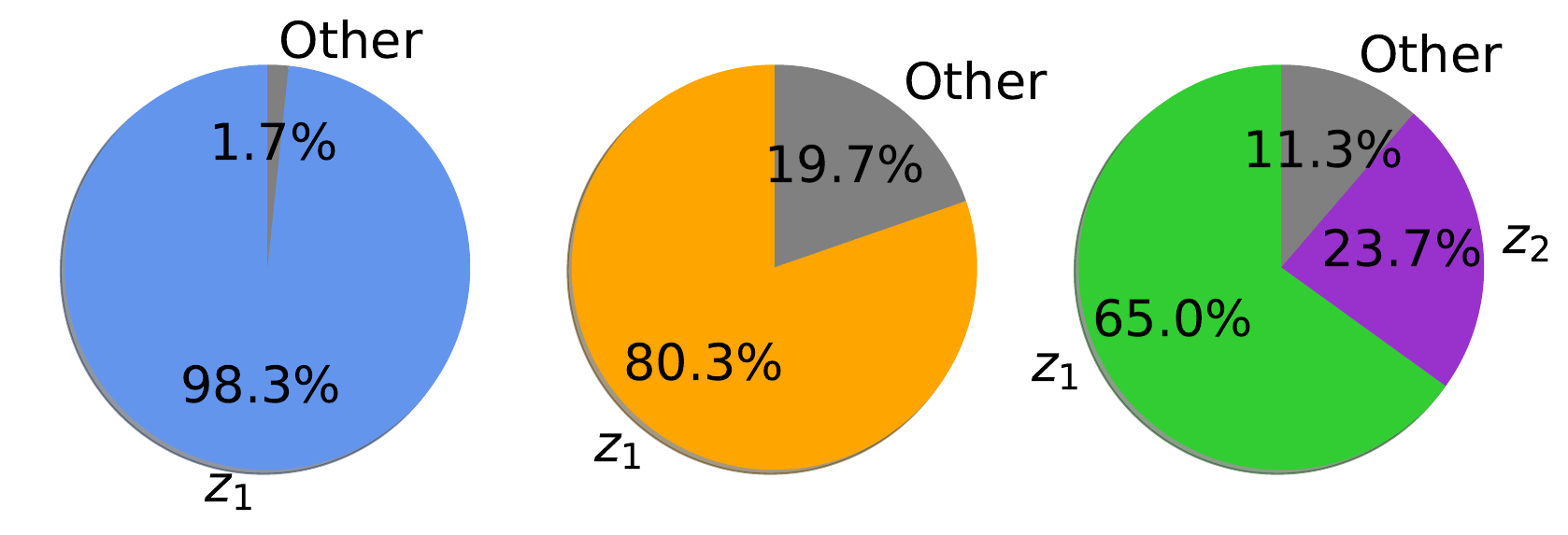}
	\end{minipage}
% \caption{Sensitivity of $Q\circ\bb{h}$ to $z_1$ as a percentage of total.}
\end{subfigure}
\end{center}
\vspace{0.3pc}
\begin{subfigure}[b]{1\textwidth}
\centering
\begin{minipage}{.04\textwidth}
    (c) 
\end{minipage}
    \begin{minipage}{.95\textwidth}
\includegraphics[width=\textwidth]{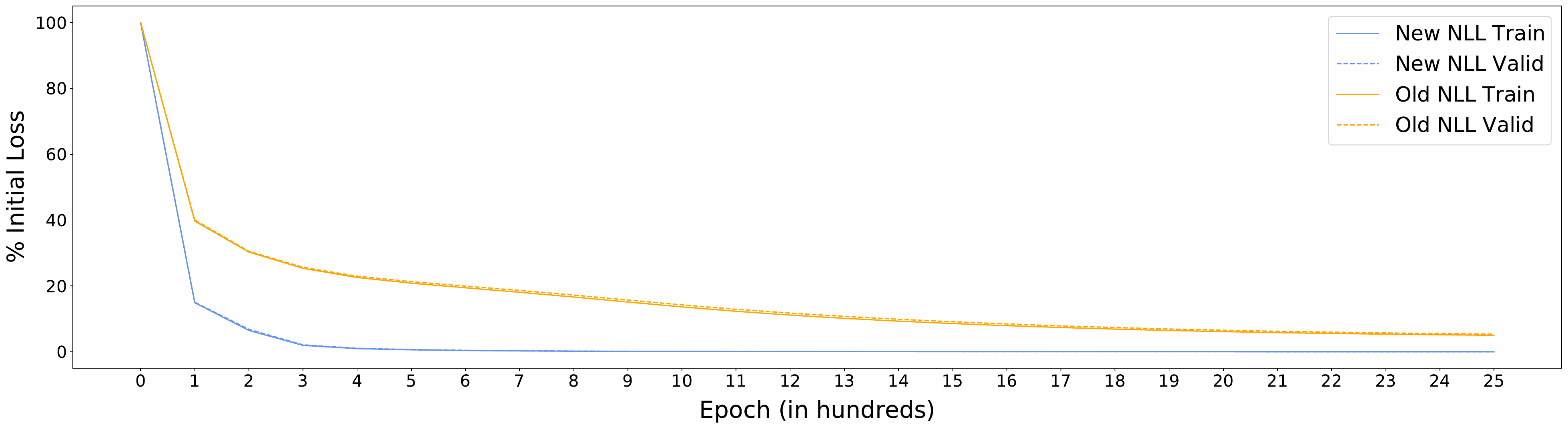}
\end{minipage}	
% \caption{NN regression: New NLL, Old NLL, 1-D AS, 2-D AS.}
\end{subfigure}
\caption{(a) Relative regression errors on $K\circ\bb{h}$; (b) Relative sensitivity of $K\circ\bb{h}$ to $\bb{z}_A$ as a percentage of total; (c) Relative value of loss during the first 2500 epochs of NLL training.}
\label{fig:burgersmore}
\end{figure}

\begin{figure}[!htb]
    \centering
	\begin{minipage}{.04\textwidth}
    (a) 
    \end{minipage}
    \,\,
    \begin{minipage}{.36\textwidth}
    \includegraphics[width=0.80\textwidth]{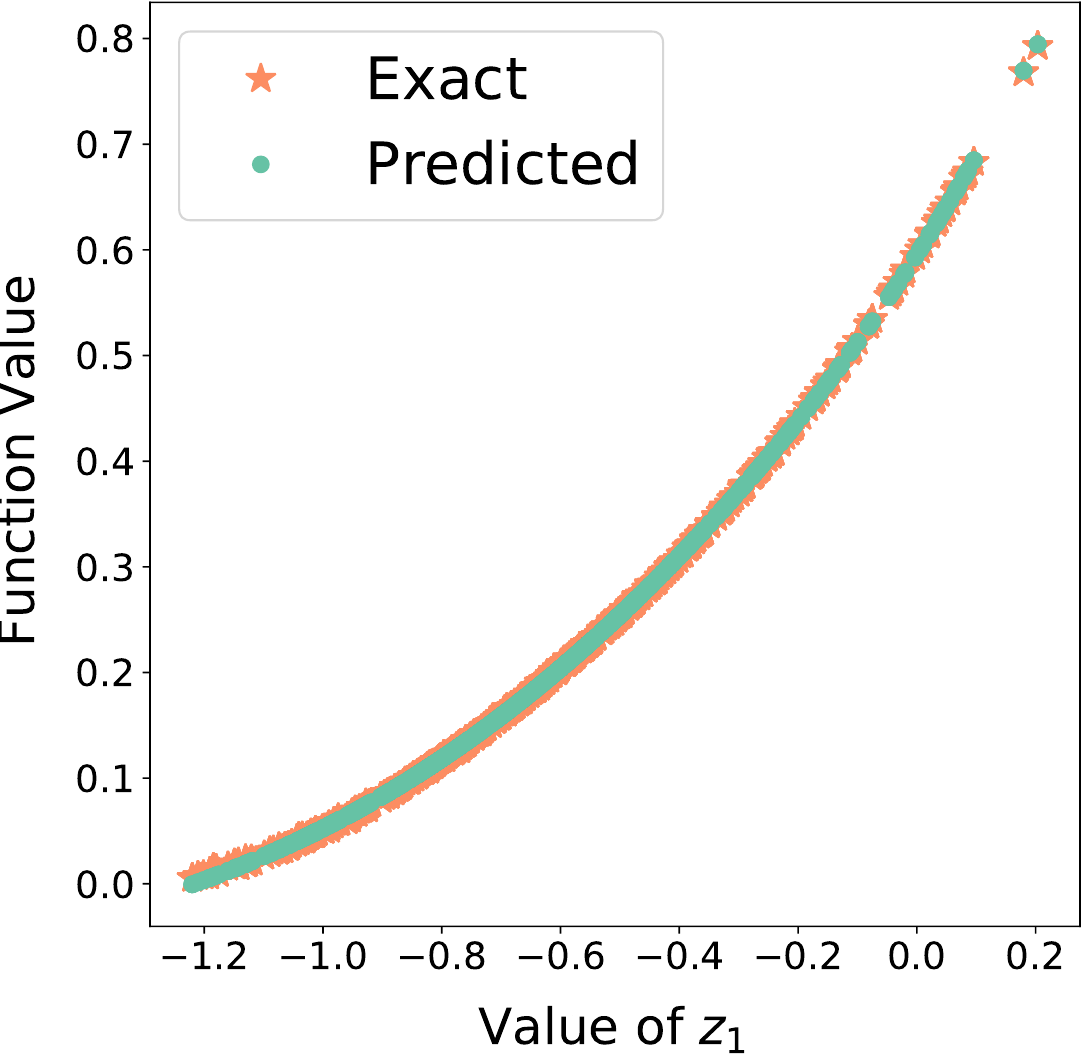}
    \end{minipage}
    \begin{minipage}{.04\textwidth}
    (b) 
    \end{minipage}
    \,\,
    \begin{minipage}{.36\textwidth}
    \includegraphics[width=0.80\textwidth]{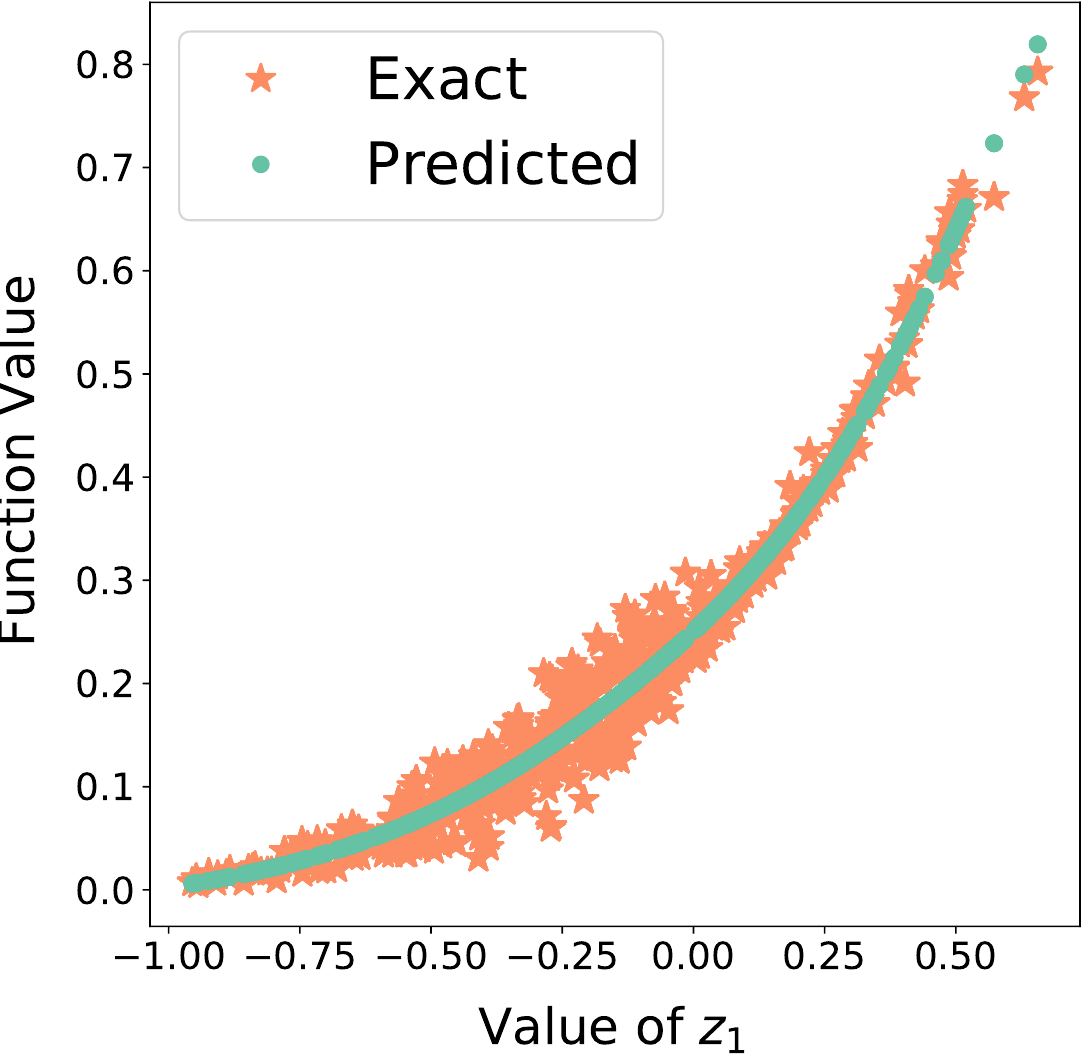}
    \end{minipage}
    
    \begin{minipage}{.04\textwidth}
    (c) 
    \end{minipage}
    \,\,
    \begin{minipage}{.36\textwidth}
    \includegraphics[width=0.80\textwidth, valign=t]{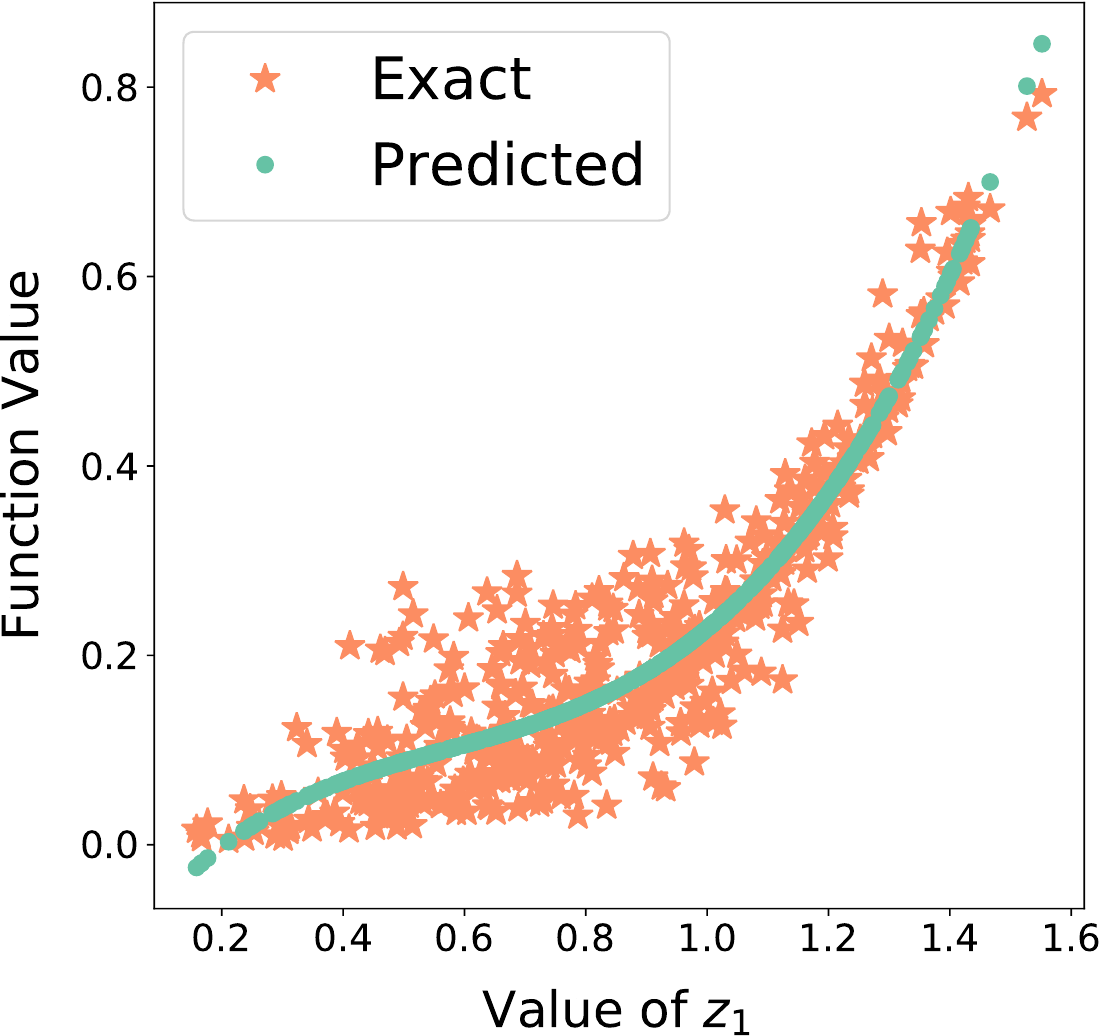}
    \end{minipage}
    \begin{minipage}{.04\textwidth}
    (d) 
    \end{minipage}
    \,\,
    \begin{minipage}{.36\textwidth}
    \includegraphics[width=0.80\textwidth, valign=t]{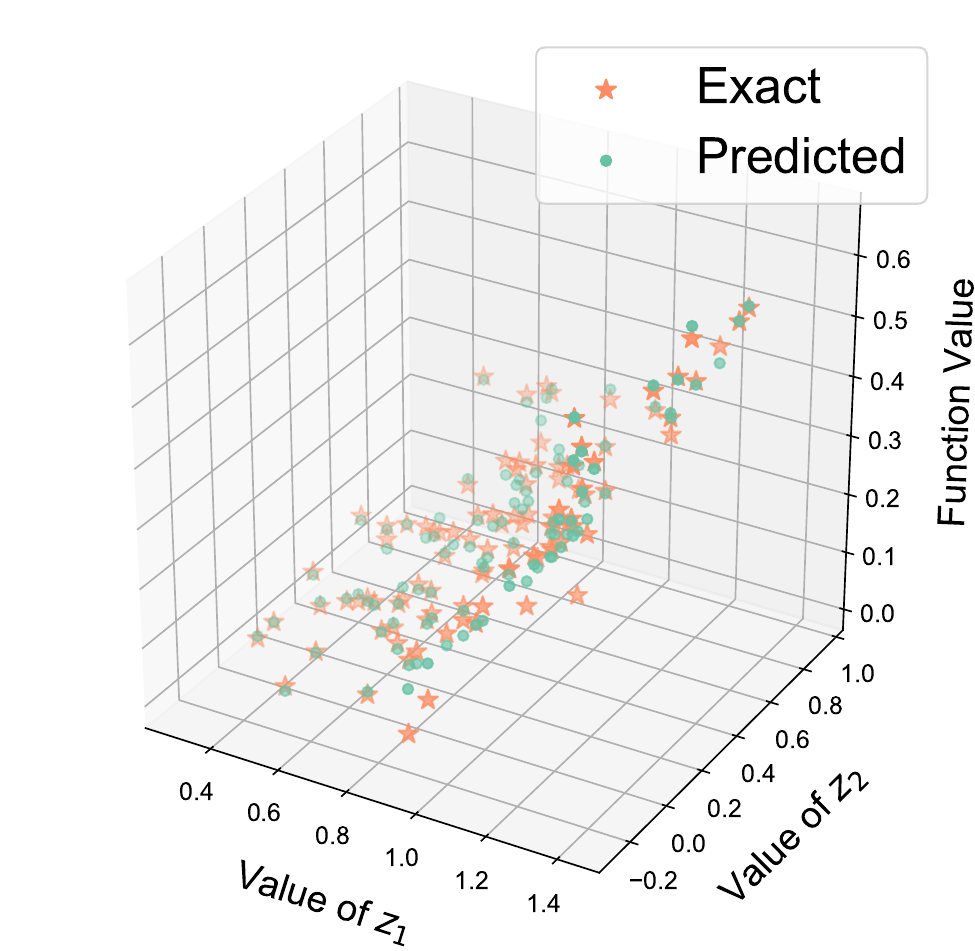}
    \end{minipage}
    \caption{Regression on $K\circ \bb{h}$: (a) New NLL; (b) Old NLL; (c) 1-D AS; (d) 2-D AS.}
    \label{fig:burgersreg}
\end{figure}

Results are displayed in Figures~\ref{fig:burgersmore} and \ref{fig:burgersreg}.  Predictably, both versions of NLL are able to reproduce a reasonable one-dimensional approximation, although the New NLL approximation is much more accurate.  Moreover, while a one-dimensional AS approximation can only capture the general trend in the function, a two-dimensional AS approximation  (slightly) outperforms the Old NLL approximation.  When using New NLL,  Figure~\ref{fig:burgersmore} (a) and (b) show that 100 training samples is sufficient for a sensitivity concentration of 98\% and regression errors of around 0.5\%.

% \begin{figure}[!htb]
%     \centering
%     % \begin{minipage}{\textwidth}
%     % \centering
%     \includegraphics[width=0.30\textwidth]{figs/mine_burgers2500.pdf}
%     \hspace{1.5pc}
%     \includegraphics[width=0.30\textwidth]{figs/theirs_burgers2500.pdf}
%     % \end{minipage}
%     \begin{minipage}{\textwidth}
%     \centering
%     \includegraphics[width=0.30\textwidth, valign=t]{figs/burgers_1dAS.pdf}
%     \hspace{1.5pc}
%     \includegraphics[width=0.30\textwidth, valign=t]{figs/burgers_2dASnew.pdf}
%     \end{minipage}
%     \caption{NN regression: New NLL, Old NLL, 1-D AS, 2-D AS.}
% \end{figure}

% \begin{figure}[h]
% \begin{center}
% \begin{center}
% \begin{subfigure}[b]{0.5\textwidth}
% \includegraphics[width=\textwidth]{figs/burgersBars.pdf}
% \caption{Relative errors of the low-dimensional regression.}
% \end{subfigure}%
% \begin{subfigure}[b]{0.5\textwidth}
% \vspace{-0.5pc}
% \includegraphics[width=1\textwidth]{figs/burgersPie.pdf}
% \vspace{0.5pc}
% \caption{Sensitivity of $R_0\circ\bb{h}$ to $z_1$ as a percentage of total.}
% \end{subfigure}
% \end{center}
% \vspace{0.3pc}
% \begin{subfigure}[b]{1\textwidth}
% \centering
% \includegraphics[width=0.23\textwidth]{figs/mine_burgers2500.pdf}
% \includegraphics[width=0.23\textwidth]{figs/theirs_burgers2500.pdf}
% \includegraphics[width=0.22\textwidth]{figs/burgers_1dAS.pdf}
% \includegraphics[width=0.24\textwidth]{figs/burgers_2dASnew.pdf}
% \caption{NN regression: New NLL, Old NLL, 1-D AS, 2-D AS.}
% \end{subfigure}
% \end{center}
% \caption{Results of experiment on $Q_2$ with 2500 training data. }
% \label{fig:burgers}
% \end{figure}

\section{Conclusion}
An improved version of the NLL algorithm from \cite{NLL} has been proposed which reduces the input dimension to one in every case.  By reformulating the central learning problem as the minimization of a Dirichlet-type energy functional, good stability and convergence properties are exhibited despite the use of sparse and high-dimensional training data.  Through various illustrative examples it has been demonstrated that New NLL has several benefits over the original NLL algorithm and the linear method of Active Subspaces, including faster training than Old NLL and a sharper, more complete dimension reduction.  Results show that regression approximations trained after New NLL are also more accurate, leading to confident prediction when approximating functionals of ODE/PDE solutions.  Future work includes obtaining rigorous estimates on the data dependence and algorithmic convergence of New NLL, as well as studying the connection between harmonic maps and minimizers of the NLL algorithm. Finally, it remains to investigate the performance of NLL when the level sets of $f$ in $U$ cannot be covered by just one coordinate chart.  Does a procedure motivated by the local IFT still produce a good mapping in this case?  If not, is there a straightforward modification which yields better results?  Although these concerns can always be addressed in principle by shrinking the domain $U$, it would be interesting to have an algorithm which is agnostic to such topological considerations.

%%%% Acknowledgments %%%%%%%%
\section*{Acknowledgments}
This work is partially supported by U.S. Department of Energy  Scientific Discovery through Advanced Computing under grants DE-SC0020270 and
DE-SC0020418.  The authors are grateful to the anonymous referee for helpful suggestions regarding the presentation.

%%%% Bibliography  %%%%%%%%%%

\bibliographystyle{unsrt} 
\bibliography{AMLbib}





% \end{thebibliography}
\end{document}